\pdfoutput=1
\documentclass[11pt, a4paper]{article}

\usepackage[final]{acl2022}

\usepackage{times}
\usepackage{latexsym}

\usepackage[T1]{fontenc}
\usepackage[utf8]{inputenc}

\usepackage{microtype}

\usepackage{amsmath}
\usepackage{lipsum}% http://ctan.org/pkg/lipsum
\usepackage{graphicx}% http://ctan.org/pkg/graphicx

\usepackage[english]{babel}
\usepackage{algorithm}
\usepackage[noend]{algpseudocode}
\makeatletter
\algnewcommand{\LineComment}[1]{\Statex \hskip\ALG@thistlm \(\triangleright\) #1}
\makeatother

\usepackage{xcolor}
\usepackage{multicol}

\usepackage{appendix}
\usepackage{etoolbox}

\BeforeBeginEnvironment{appendices}{\clearpage}
\title{Rare Tokens Degenerate All Tokens: \\ Improving Neural Text Generation via Adaptive Gradient Gating \\ for Rare Token Embeddings}

\author{Sangwon Yu$^1$   Jongyoon Song$^1$    Heeseung Kim$^1$    Seong-min Lee$^3$ \\ {\bf Woo-Jong Ryu$^3$}  {\bf    Sungroh Yoon$^{1,2,}$\Thanks{\ Corresponding author.}} \\
   $^1$Data Science \& AI Laboratory, Seoul National University, Korea \\
   $^2$ASRI, ECE, GSAI, and INMC, Seoul National University, Korea \\
   $^3$AIRS Company, Hyundai Motor Group, Korea \\
   \{\texttt{dbtkddnjs96}, \texttt{coms1580}, \texttt{gmltmd789}, \texttt{sryoon}\}\texttt{@snu.ac.kr} \\
   \{\texttt{blueworm7}, \texttt{woojong.ryu}\}\texttt{@hyundai.com} \\ }

\begin{document}
\maketitle
\begin{abstract}
Recent studies have determined that the learned token embeddings of large-scale neural language models are degenerated to be anisotropic with a narrow-cone shape. This phenomenon, called the representation degeneration problem, facilitates an increase in the overall similarity between token embeddings that negatively affect the performance of the models. Although the existing methods that address the degeneration problem based on observations of the phenomenon triggered by the problem improves the performance of the text generation, the training dynamics of token embeddings behind the degeneration problem are still not explored. In this study, we analyze the training dynamics of the token embeddings focusing on rare token embedding. We demonstrate that the specific part of the gradient for rare token embeddings is the key cause of the degeneration problem for all tokens during training stage. Based on the analysis, we propose a novel method called, \textit{adaptive gradient gating} (AGG). AGG addresses the degeneration problem by gating the specific part of the gradient for rare token embeddings. Experimental results from language modeling, word similarity, and machine translation tasks quantitatively and qualitatively verify the effectiveness of AGG.   
\end{abstract}

\section{Introduction}
Neural language models have been developed with various architectures during recent years (\citealp{Graves2013GeneratingSW}; \citealp{Bahdanau2015NeuralMT}; \citealp{Gehring2017ConvolutionalST}; \citealp{Vaswani2017AttentionIA}). Despite the improvement in model architectures, models usually share the same process for input and output. They process token embeddings as inputs to compute contextualized features and subsequently project the features into a categorical distribution of tokens at the output softmax layer whose weight is token embedding matrix (\citealp{Merity2017PointerSM}; \citealp{Yang2018BreakingTS}; \citealp{Press2017UsingTO}). Recent studies have determined that the learned embedding distribution is biased in a common direction, thereby resulting in a narrow cone-shaped anisotropy (\citealp{Mu2018AllbuttheTopSA}; \citealp{ethayarajh-2019-contextual}; \citealp{Gao2019RepresentationDP}; \citealp{bis2021tmic}). This phenomenon, named the representation degeneration problem by \citet{Gao2019RepresentationDP}, increases the overall similarity between embeddings, and leads to a problem in which the expressiveness of the token embeddings decreases. Therefore, it is difficult for the model to learn the semantic relationship between the tokens and to generate high quality texts. Existing studies addressing this problem suggest methods that apply post-processing or regularization techniques to all token embeddings based on the observed phenomena owing to the degeneration problem (\citealp{Mu2018AllbuttheTopSA}; \citealp{Gao2019RepresentationDP}; \citealp{pmlr-v97-wang19f}; \citealp{Wang2020ImprovingNL}; \citealp{bis2021tmic}). Although these works improve the quality of token embeddings and generated texts, it is still not clear how token embeddings become degenerate during training procedure. Also, there exists the problem of over regularization for the token embeddings whose semantic relationships are trained well because the above methods are applied for all token embeddings.
\begin{figure*}[t]
     \centering
     \begin{subfigure}[t]{0.24\textwidth}
         \centering
         \includegraphics[width=\textwidth]{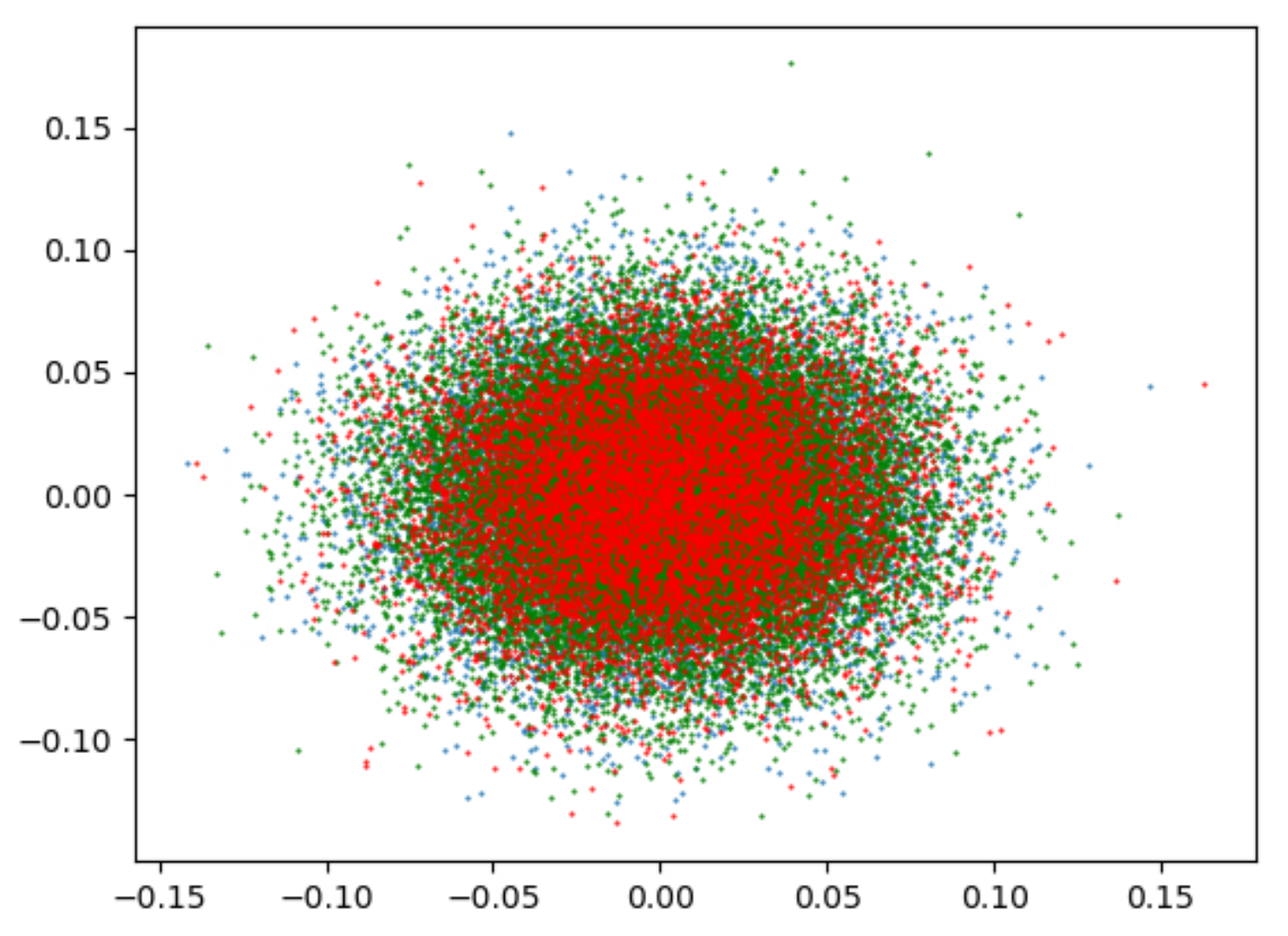}
         \caption{Training step 100}
         \label{fig:y equals x}
     \end{subfigure}
     \hfill
     \begin{subfigure}[t]{0.24\textwidth}
         \centering
         \includegraphics[width=\textwidth]{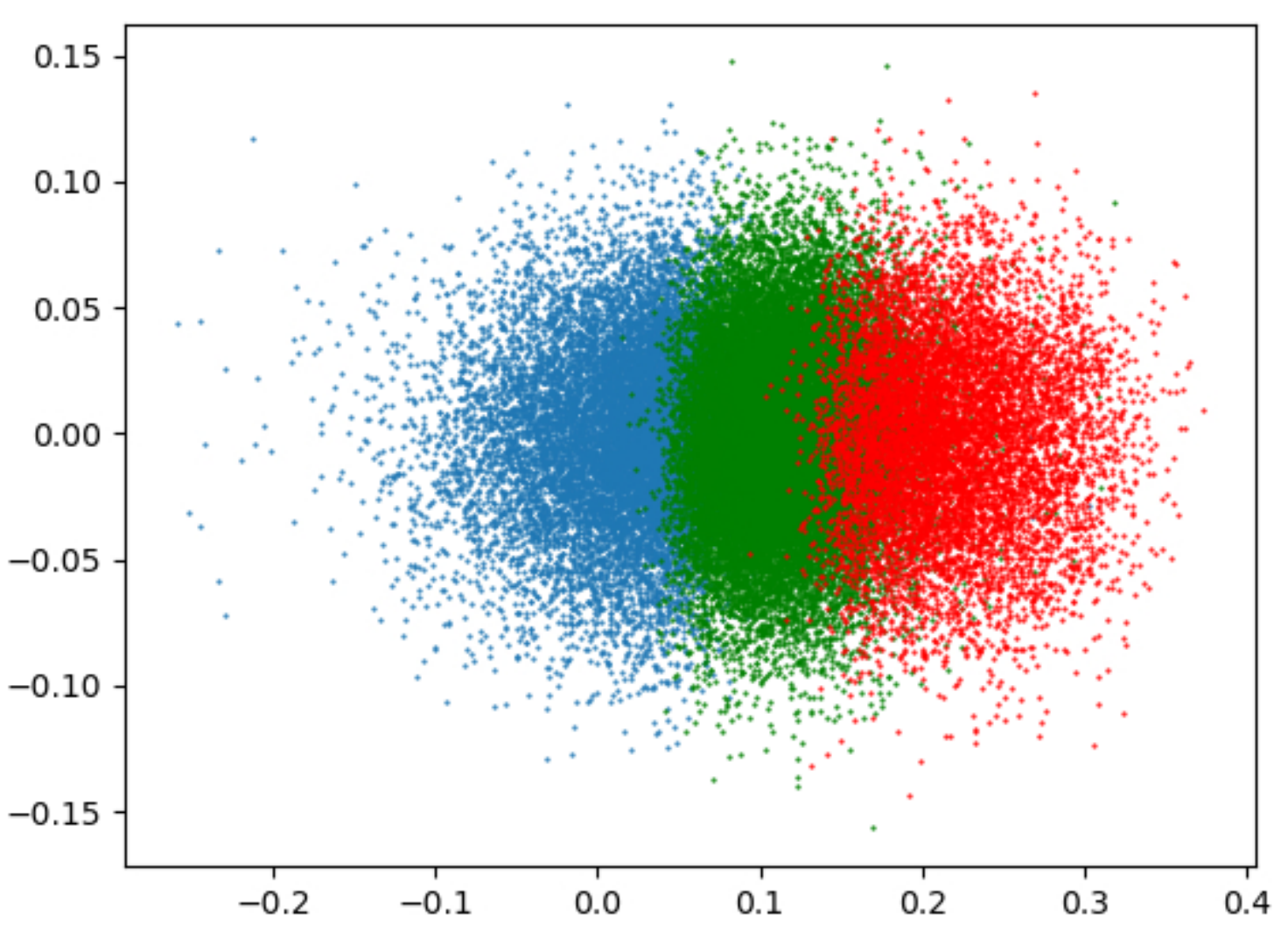}
         \caption{Training step 500}
         \label{fig:three sin x}
     \end{subfigure}
     \hfill
     \begin{subfigure}[t]{0.24\textwidth}
         \centering
         \includegraphics[width=\textwidth]{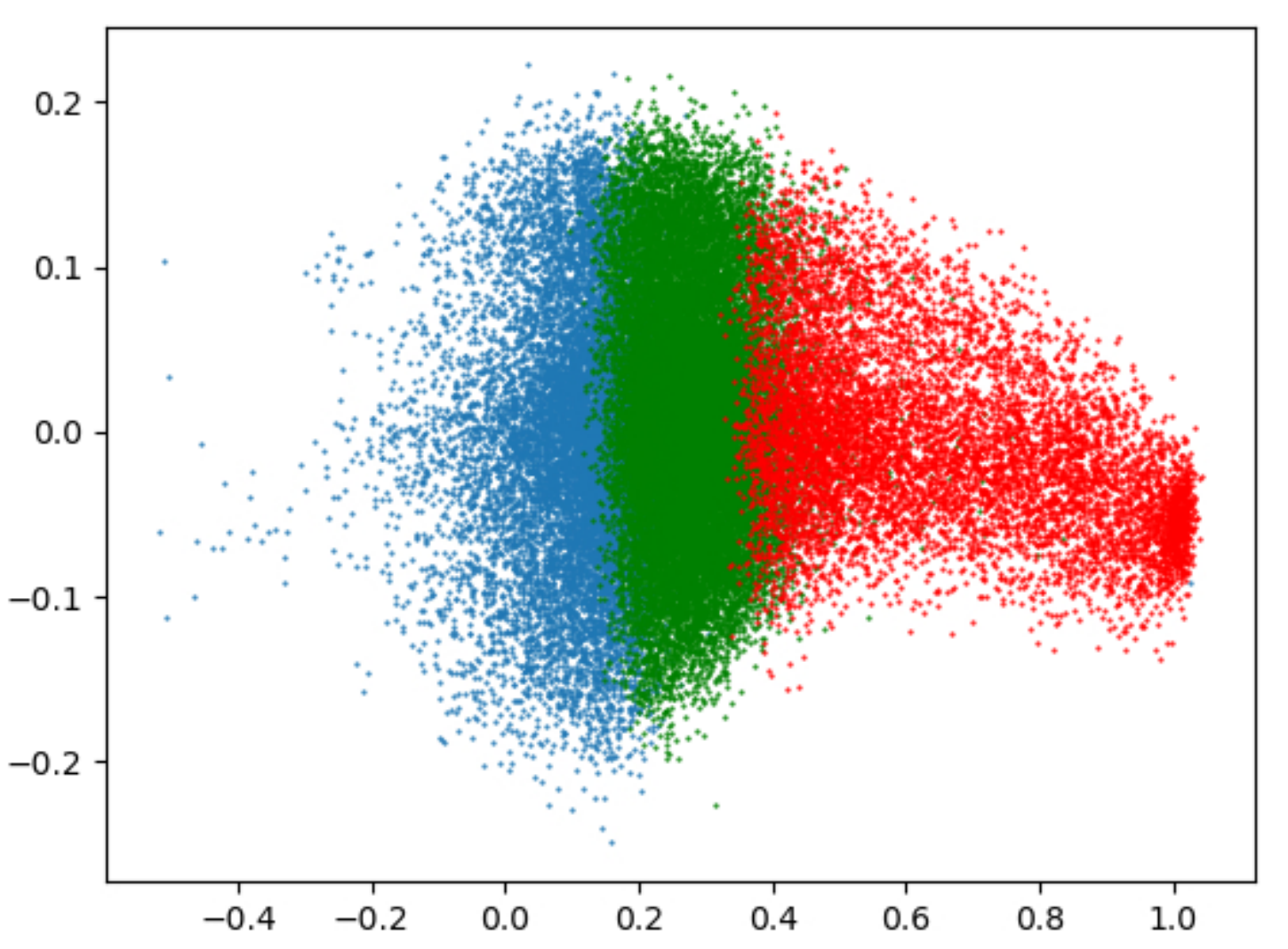}
         \caption{Training step 1500}
         \label{fig:five over x}
     \end{subfigure}
     \hfill
     \begin{subfigure}[t]{0.24\textwidth}
         \centering
         \includegraphics[width=\textwidth]{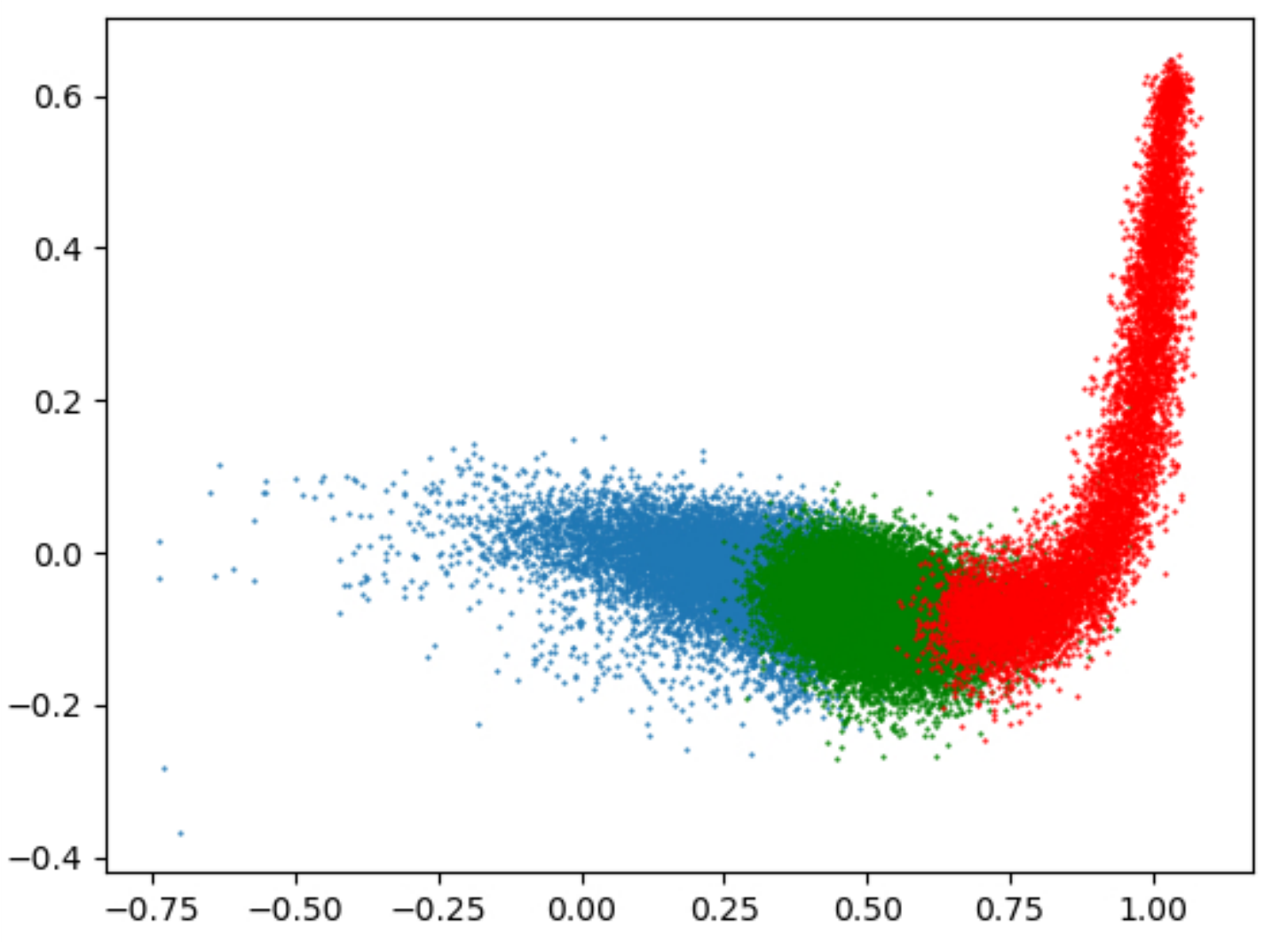}
         \caption{Training step 3500}
         \label{fig:five over x}
     \end{subfigure}
        \caption{Visualization of token embeddings of language model trained on WikiText-103. Red, green, and blue points represent rare, medium, and frequent groups respecively. (a), (b), (c), (d) present a visualization of each training step.}
        \label{fig_trainingsch}
\end{figure*}

In this study, we conduct empirical studies about training dynamics of token embeddings, focusing on rare token embeddings. By observing the initial training dynamics of token embeddings grouped based on appearance frequency, we hypothesize that the degeneration of the rare token embeddings triggers the degeneration of the embeddings of the remaining tokens. We show that the entire degeneration problem is mitigated by only freezing rare tokens during training, and we demonstrate that the main cause of the entire degeneration problem is the specific part of the gradient for rare token embeddings. This gradient part pushes away rare token embeddings from the feature vector of the non-rare targets in the current training sample. Based on the analysis, we propose a new method, \textit{adaptive gradient gating} (AGG). With a dynamic grouping of rare tokens at each training step, AGG solves the entire degeneration problem by gating a specific part of the gradient that is solely about rare tokens. Because AGG is optimized to target the main cause of the degeneration problem, rare token embeddings, it can prevent the over regularization problem about frequent token embeddings which occurs in other methods addressing the degeneration problem. The proposed method is evaluated in three tasks: language modeling, word similarity, and machine translation. The AGG outperforms the baseline and other existing methods in all tasks. In addition, it shows compatibility with other method that addresses the neural text degeneration problem. Via qualitative studies, we identify a correlation between our method and the frequency bias problem of learned embeddings (\citealp{Gong2018FRAGEFW}; \citealp{Ott2018AnalyzingUI}).

\section{Background}

\subsection{Text Generation of Neural Language Models}
Neural language generative models process text generation tasks as conditional language modeling, in which the model is typically trained by minimizing the negative log likelihood of the training data. With a vocabulary of tokens $V = \{v_1, ..., v_N\}$ and embedding vectors $\{\textbf{w}_1, ..., \textbf{w}_N\}$, where $\textbf{w}_i$ corresponds to token $v_i$, at every training step, the model obtains a mini-batch input and target text corpus pair ($\textbf{x}$, $\textbf{y}$), where $x_i$, $y_i\in V$, and $\textbf{y}\in V^T$. The conditional probability for the target token $y_t$, $P_\theta(y_t|\textbf{h}_t)$, where $\textbf{h}_t$ is a context feature vector of the $t$-th position of the generated text conditioned by ($\textbf{x}$, $y_{<t}$), and $\theta$ denotes model parameters, which is defined as follows.
\begin{equation}\label{eq1}
    P_\theta(y_t|\textbf{h}_t)=\frac{\exp{(\textbf{h}_t\textbf{w}^T_{I(y_t)}})}{\sum_{l=1}^{N}\exp{(\textbf{h}_t\textbf{w}^T_l})},
\end{equation}
where $\textbf{w}$ is the output token embedding which roles the weight of the output softmax layer, and $I(y_t)$ represents the index of token $y_t$. The negative log likelihood loss for an input and target pair ($\textbf{x}$, $\textbf{y}$), $L_{NLL}$ is expressed as follows.
\begin{equation}\label{eq2}
        L_{NLL} = -\sum_{t=1}^{T}\log{P_\theta(y_t|\textbf{h}_t)}. 
\end{equation}
% Input $\textbf{x}$ is task-specific, \textit{e.g.,} prefix at text continuation, source sentence at machine translation, and input image at image captioning.

\begin{table*}
\centering
\begin{tabular}{l||cccc|cccc}
\hline
\toprule
\multirow{2}{*}{\textbf{Methods}} &
\multicolumn{4}{c|}{\textbf{PPL} $\downarrow$} & 
\multicolumn{4}{c}{$\mathbf{I(W)}$ $\uparrow$}  \\
 & Freq & Med & Rare & Total & Freq & Med & Rare & Total \\
\hline
MLE & 16.58 & 224.24 & 813.76 & 20.77 & 0.426 & 0.286 & 0.198 & 0.293 \\
\hline
Freeze & 16.48 & 233.92 & 3017.53 & 20.78 & 0.840 & 0.651 & 0.831 & 0.739 \\
\bottomrule
\end{tabular}
\caption{Perplexity and $I(\textbf{W})$ for each token groups. Lower is better for PPL and higher is better for $I(\textbf{W})$.}
\label{table_freezeall}
\end{table*}

\begin{figure*}[t]
     \centering
     \begin{subfigure}[t]{0.32\textwidth}
         \centering
         \includegraphics[width=\textwidth]{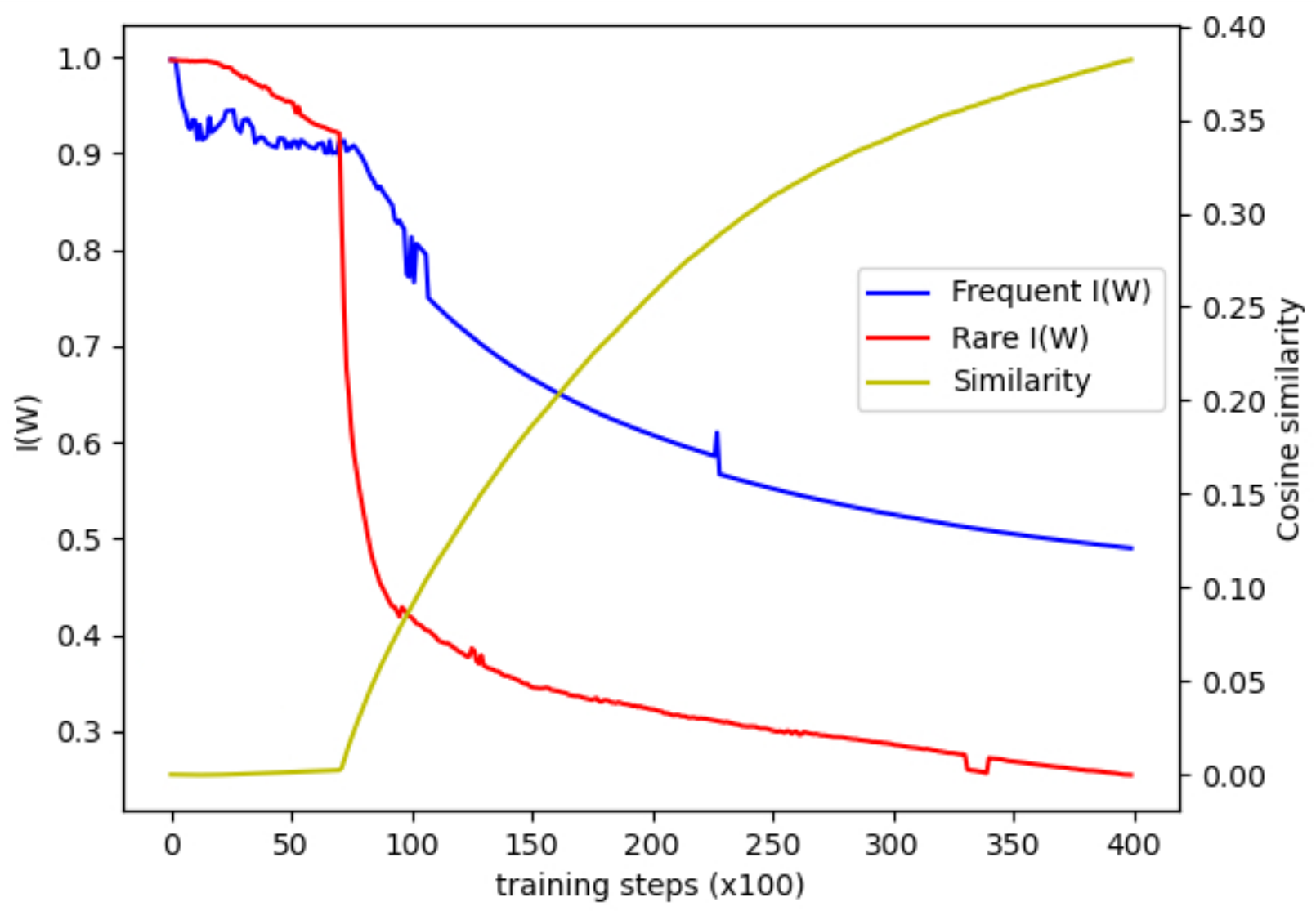}
         \caption{freeze until step 7k}
     \end{subfigure}
     \hfill
     \begin{subfigure}[t]{0.32\textwidth}
         \centering
         \includegraphics[width=\textwidth]{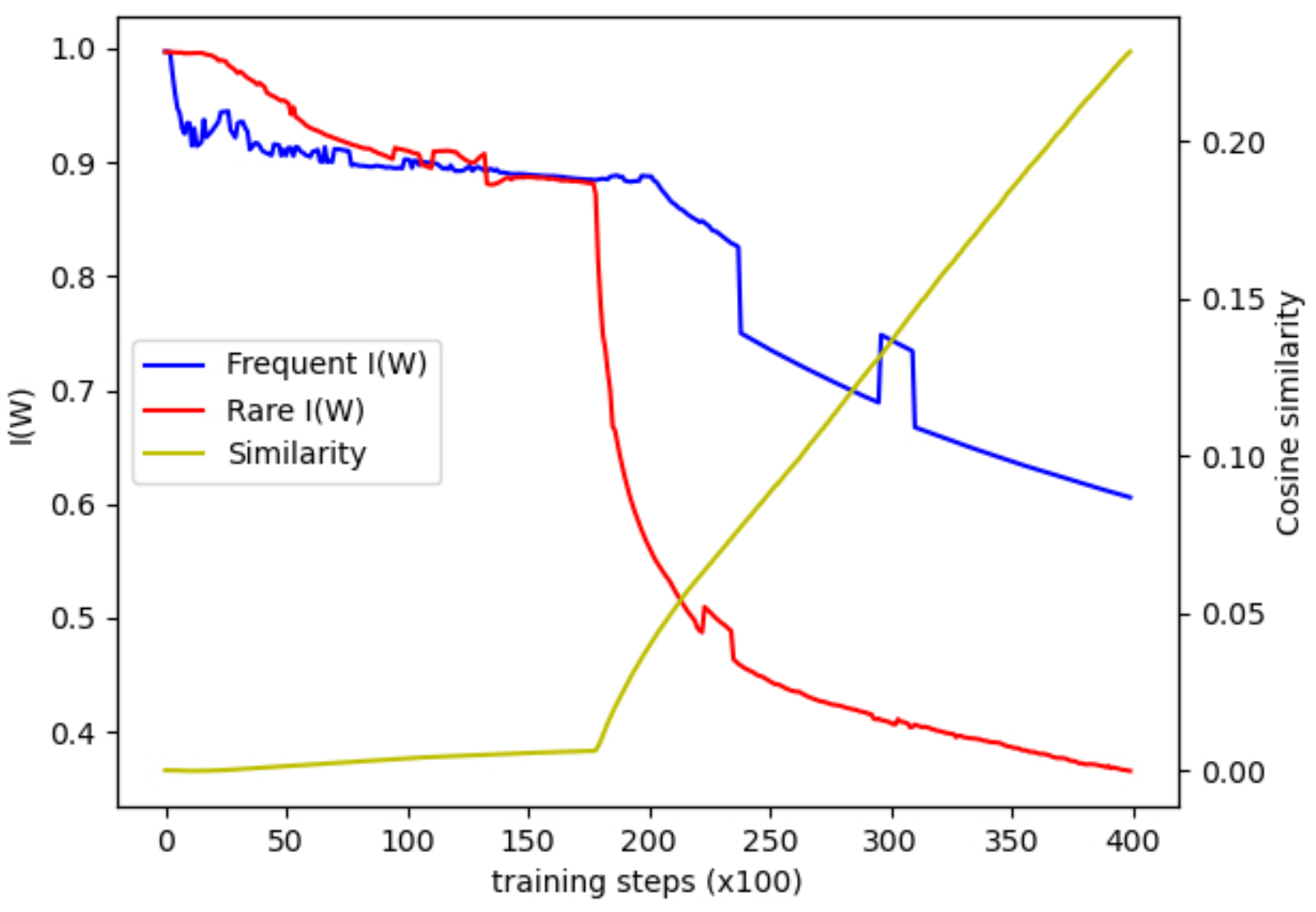}
         \caption{freeze until step 18k}
     \end{subfigure}
     \hfill
     \begin{subfigure}[t]{0.32\textwidth}
         \centering
         \includegraphics[width=\textwidth]{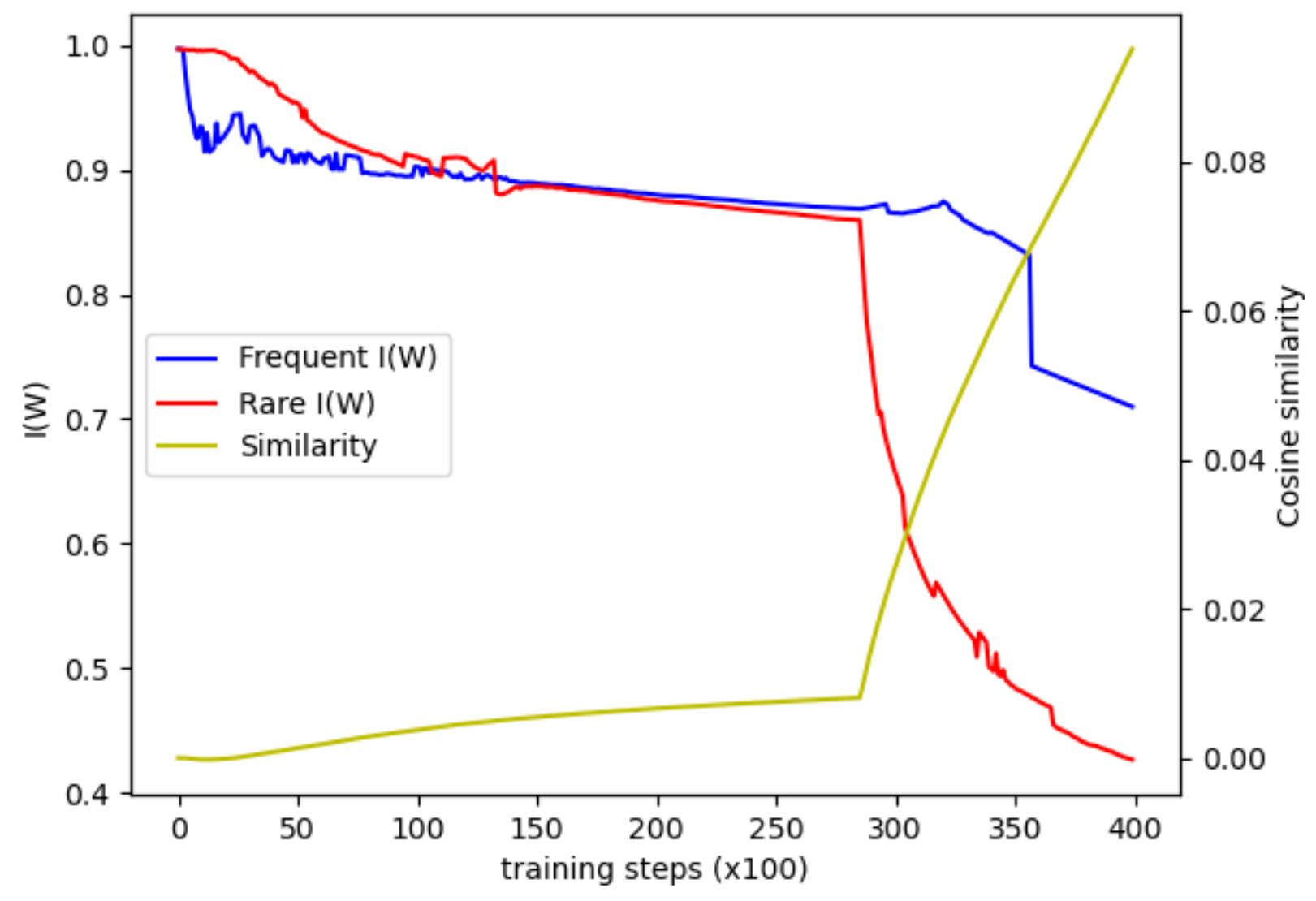}
         \caption{freeze until step 29k}
     \end{subfigure}
     \caption{Plot of $I(\textbf{W})$ for rare and frequent groups and average cosine similarity between rare and frequent embeddings when freezing the training of rare tokens until specific training steps.}
     \label{fig_melt}
\end{figure*}

\subsection{Embedding Problems in Neural Language Models}
Recent studies on the geometric properties of contextual embedding space have observed that the distribution of embedding vectors is far from isotropic and occupies a relatively narrow cone space(\citealp{Mu2018AllbuttheTopSA}; \citealp{Liu2019UnsupervisedPO}; \citealp{ijcai2019-761}; \citealp{ethayarajh-2019-contextual};). \citet{Gao2019RepresentationDP} named this phenomenon the \textit{representation degeneration problem}. This degeneration problem results in an increase in the overall cosine similarity between token embeddings, making it difficult for the model to learn semantic relationships between tokens. \citet{2020StolenProb} demonstrated that the norm information of the token embeddings is so dominant that angle information about the feature vector is ignored when calculating the logits in the output layer. Owing to this structural weakness of the embedding space, embeddings with small norms are always assigned with a low probability, which reduces the diversity of the text generated by the model. Anisotropy of the embedding space is a still problem for the pre-trained large language models, and language models with improved isotropic embedding space performs well in downstream tasks(\citealp{bis2021tmic}; \citealp{Rajaee2021ACA}). 

Although the problem has been theoretically analyzed in several studies, existing methods are based on the observed phenomena as a result of the problem. To mitigate the phenomena observed from the problem, the post-processing of the embedding vectors(\citealp{Mu2018AllbuttheTopSA}; \citealp{bis2021tmic}) or regularization terms about the phenomena(\citealp{Gao2019RepresentationDP}; \citealp{pmlr-v97-wang19f}; \citealp{Wang2020ImprovingNL}; \citealp{Zhang2020RevisitingRD}) were introduced. These methods are applied to all token embeddings, so there is the problem of over regularization for the embeddings whose semantic relationship is trained well. Also, methodologies based on the training dynamics of the token embeddings concerning the degeneration problem remain subject to study.

Frequency bias in embedding space is another problem. \citet{Ott2018AnalyzingUI} conducted a comprehensive study on the under-estimation of rare tokens in neural machine translation. \citet{Gong2018FRAGEFW} observed that embeddings in the language model were biased towards frequency and proposed an adversarial training scheme to address this problem.

\section{Empirical Study: Token Embedding Training Dynamics led by Rare Tokens} \label{sec3}

\begin{table*}
\centering
\begin{tabular}{l||cccc|cccc}
\hline
\toprule
\multirow{2}{*}{\textbf{Methods}} &
\multicolumn{4}{c|}{\textbf{PPL} $\downarrow$} & 
\multicolumn{4}{c}{$\mathbf{I(W)}$ $\uparrow$} \\
 & Freq & Med & Rare & Total & Freq & Med & Rare & Total \\
\hline
MLE & 16.58 & 224.24 & 813.76 & 20.77 & 0.426 & 0.286 & 0.198 & 0.293 \\
\hline
Freeze (b) \& (c) & 17.41 & 247.89 & 66.41 & 21.79 & 0.323 & 0.693 & 0.551 & 0.536 \\

Freeze (b) & 16.99 & 240.72 & 65.76 & 21.26 & 0.495 & 0.561 & 0.678 & 0.748 \\

Freeze (c) & 16.61 & 220.07 & 645.24 & 20.76 & 0.443 & 0.276 & 0.15 & 0.317 \\

\bottomrule
\end{tabular}
\caption{Perplexity and $I(\textbf{W})$ for each token group at gradient partial freezing experiment.}
\label{table_freezepart}
\end{table*}

\subsection{Initial Training Dynamics of Embeddings}
To analyze the training procedure of token embeddings, we train a Transformer language model at the WikiText-103 dataset from scratch. Whole vocabulary tokens are divided into three groups: frequent, medium, and rare groups. Based on the appearance frequency in the training corpus, the 30\%, 50\%, and 20\% tokens are assigned to the frequent, medium, and rare group. We visualize the initial training dynamics of these groups via the projection of the embeddings into 2D, using singular value decomposition (SVD) projection. As illustrated in Figure \ref{fig_trainingsch}, rare groups degenerate first, as they emerge from the entire embedding distribution. Subsequently, other groups also start to degenerate, following the degeneration of the rare group. Based on this observation, we hypothesize that \textit{the degeneration of rare token embeddings induces the degeneration of non-rare token embeddings}.

\subsection{Rare Tokens Degenerate Non-Rare Tokens}
Because Transformer \citep{Vaswani2017AttentionIA} is representative of the current language models, we adopt the 6-layer Transformer decoder model architecture for an empirical study on the training dynamics of embedding vectors. The model is trained in language modeling task using WikiText-103 dataset \citep{Merity2018RegularizingAO}. Experimental details regarding the model and training hyperparameter configurations can be found in the Appendix \ref{appenB}. To verify the hypothesis of the previous subsection, we train a model while freezing the rare group token embeddings in their initial states during training, and compare it to the baseline model, where all embeddings are trained with negative log-likelihood loss. In addition, we train the models of various settings relative to freezing steps and examine whether the degeneration of rare token embeddings depends on when training of rare embeddings begins. 

The performance of the models is evaluated in two ways; the likelihood and isotropy of token embeddings. Perplexity \citep{10.5555/944919.944966} is adopted to evaluate the performance of the likelihood of the model. To measure the isotropy of the token embedding distribution, we adopt the partition function $Z(\textbf{a})=\sum_{i=1}^{N}\exp{(\textbf{w}_i\textbf{a}^T)}$ defined in \citet{arora-etal-2016-latent}, where $\textbf{w}_i$ denotes the embedding vector of token $v_i$, and $\textbf{a}$ represents a unit vector. Lemma 2.1. in \citet{arora-etal-2016-latent} demonstrate that if the embedding vectors are isotropic, $Z(\textbf{a)}$ is approximately constant. Based on this property, we measure the isotropy of an embedding matrix $\textbf{W}$ using $I(\textbf{W})$, which is defined as follows.
\begin{equation}\label{eq3}
    I(\textbf{W})=\frac{\min_{\textbf{a}\in \textbf{X}}Z(\textbf{a})}{\max_{\textbf{a}\in \textbf{X}}Z(\textbf{a})},
\end{equation}
where $I(\textbf{W})\in [0,1]$ and $\textbf{X}$ represents the set of eigenvectors of $\textbf{W}^T\textbf{W}$ (\citealp{Mu2018AllbuttheTopSA}; \citealp{Wang2020ImprovingNL}; \citealp{bis2021tmic}).  Furthermore, we measure the relatedness between the rare and frequent group token embeddings to verify that the degeneration of the frequent group follows the degeneration of the rare group. We calculate the average cosine similarity between the rare and frequent group embeddings to measure the relatedness.

Table \ref{table_freezeall} shows the comparison of the baseline model and the model with frozen rare tokens. We denote the baseline as "MLE" and the freezing method as "Freeze". Surprisingly, the PPL of frequent group tokens and overall $I(\textbf{W})$ improved by simply not training the rare token embeddings. Figure \ref{fig_melt} illustrates the change in $I(\textbf{W})$ for the frequent and rare token embeddings, including the similarity between frequent and rare token embeddings at various freezing step settings. Whenever the rare token embeddings start to be trained, their $I(\textbf{W})$ decreases steeply, followed by decreasing $I(\textbf{W})$ of frequent embeddings and increasing similarities between the frequent and rare embeddings. From the analysis in this subsection, we demonstrate that the entire degeneration problem can be solved by solely handling just rare embeddings during the entire training procedure.

\subsection{Finding the Primary Cause of the Degeneration Problem: From the Gradient}
With $T$ context feature vectors $\textbf{h}_i$ ($i\in[1,T]$) from the training sample, the negative log-likelihood loss gradient for the rare token embedding $\textbf{w}_r$ is calculated as follows.
\begin{equation}\label{eq4}
    \begin{aligned}
        \nabla_{\textbf{w}_r}L_{NLL}
        &=\underbrace{\sum_{y_i=v_r}(p_{r|i}-1)\textbf{h}_i}_{(a)} \\
        &+ \underbrace{\sum_{y_j\notin V_r}p_{r|j}\textbf{h}_j}_{(b)}
        + \underbrace{\sum_{y_k\in V_r}p_{r|k}\textbf{h}_k}_{(c)}, 
    \end{aligned}
\end{equation}
where $y_i$ denotes the target token for $\textbf{h}_i$, $V_r$ is the rare token vocabulary group, and $p_{r|i}$ represents the conditional probability of token $v_r$ given $\textbf{h}_i$, which is calculated as $[\text{softmax}(\textbf{h}_i\textbf{W}^T)]_r$. We divide the gradient for $\textbf{w}_r$ to 3 parts in Eq. \ref{eq4}. Part (a) pulls $\textbf{w}_r$ close to the feature vectors whose target tokens are $v_r$. Part (b) pushes away $\textbf{w}_r$ from the feature vectors whose target tokens are not rare. Part (c) pushes away $\textbf{w}_r$ from the feature vectors whose target tokens are rare. As an extension of the analysis in the previous subsection, we freeze these parts of the gradient with various settings during training to identify the key cause of the degeneration problem. In other words, depending on the settings, the specific gradient parts that will not be used for embedding training is detached from the computation graph during training stage. This can be easily implemented by \texttt{detach()} function of \texttt{Pytorch} \citep{Paszke2019PyTorchAI}. All model and training configurations are the same as in the previous sections, except those to be frozen.

Table \ref{table_freezepart} presents the results of the experiments in this subsection. We freeze the parts of the gradient for the rare tokens with three settings. Because part (a) is a key component required to train the token embedding to be aligned to the target, all settings activate part (a). We notice that when part (b) is activated (solely freezing part (c)), $I(\textbf{W})$ decreases and PPL for rare tokens increases almost 10 times compared to when part (b) is frozen. Because activating part (c) is not seen to be negative for PPL and $I(\textbf{W})$, we conclude that part (b) of Eq. \ref{eq4} is the bedrock cause for the degeneration problem. From the analysis in this section, we demonstrate that \textit{the degeneration problem could be solved to a large extent by mainly addressing the part of the gradient for rare embeddings that pushes away rare token embeddings from non-rare feature vectors}.

\section{Method} \label{sec_method}

\subsection{Dynamic Rare Token Grouping}
To handle the specific part of the gradient for the rare token embeddings studied in the previous section, we need to properly group the rare tokens. A naive approach can be used to group rare tokens based on the appearance frequency of the training corpus, as described in the previous section. However, this static grouping method is suboptimal because the model is typically trained via mini-batch training. The group of rare tokens that appeared less frequently in recent batch samples is variable in the mini-batch training. Therefore, it is necessary to dynamically group rare tokens based on token appearances in recent batch samples.

To consider the token appearances in recent batch samples, we introduce the token counter memory that remembers the number of the appearances of each token during the previous $K$ training steps. For $K$ memories, [$\textbf{m}_1, ..., \textbf{m}_K$], $\textbf{m}_t\in \mathbb{R}^N$ represents the number of appearances of each token of $N$-size vocabulary at the $t$-th previous training step. Memories are set as zero vectors at the initial stage. At each training step, the token appearance, $\textbf{a}\in \mathbb{R}^N$, is calculated as the sum of all $K$ memories: $\textbf{a}=\sum_{t=1}^{K}\textbf{m}_t$. Based on $\textbf{a}$, we determine whether token $v_i$ is in the rare token group $V_r$ as follows.
\begin{equation}\label{eq5}
    \begin{aligned}
        & \frac{a_i}{K} < \alpha \Rightarrow v_i \in V_r \\
        & \frac{a_i}{K} \geq \alpha  \Rightarrow v_i \notin V_r,
    \end{aligned}
\end{equation}
where $a_i$ is the $i$-th component of $\textbf{a}$, and $\alpha$ is a hyper-parameter in our method that controls the proportion of rare tokens in the entire vocabulary. In this study, we set $K$ to the number of iteration steps during one epoch of training stage.   

\subsection{Adaptive Gradient Gating for Rare Tokens}
After dynamically grouping the rare tokens at each training step, we need to handle a specific part of the gradient for the rare token embeddings to solve the degeneration problem of all embeddings. To solely control the gradient for rare token embeddings, we introduce a \textit{gradient gating} method for a parameter $\textbf{x}$. We define $\tilde{\textbf{x}}$ as a tensor whose value is the same as \textbf{x}, but detached from the current training graph. This implies that $\tilde{\textbf{x}}$ is considered a constant, hence, gradient about $\tilde{\textbf{x}}$ does not exist. In practice, $\tilde{\textbf{x}}$ can be easily obtained from $\textbf{x}$ using the \texttt{detach()} function of \texttt{Pytorch} \citep{Paszke2019PyTorchAI}. With $\tilde{\textbf{x}}$, we can gate the gradient for $\textbf{x}$ as follows.
\begin{equation}\label{eq6}
    \begin{aligned}
         &\textbf{x}_{gated} = \textbf{g}\odot \textbf{x} + (1-\textbf{g})\odot \tilde{\textbf{x}} \\
         &\nabla_\textbf{x}f(\textbf{x}_{gated}) = \textbf{g}\odot \nabla_\textbf{x}f(\textbf{x}),
    \end{aligned}
\end{equation}
% \begin{equation}\label{eq7}
%     \begin{aligned}
%         \frac{\partial{f(\textbf{x}_{gated})}}{\partial{\textbf{x}}} = \textbf{g}\odot \frac{\partial{f(\textbf{x})}}{\partial{\textbf{x}}},
%     \end{aligned}
% \end{equation}
where $\textbf{x}_{gated}$ is a new parameter whose value is the same as $\textbf{x}$, and $\textbf{g}\in [0,1]$ is a gate tensor. When the $\textbf{x}_{gated}$ is fed to the function $f(\cdot)$ as input, the gradient for $\textbf{x}$ is gated by $\textbf{g}$.

\begin{table*}
\centering
\begin{tabular}{l||cccc|cccc|c}
\hline
\toprule
\multirow{2}{*}{\textbf{Methods}} &
\multicolumn{4}{c|}{\textbf{PPL} $\downarrow$} &
\multicolumn{4}{c|}{\textbf{Uniq} $\uparrow$} & 
\multirow{2}{*}{$\mathbf{I(W)}$$\uparrow$} \\ 
 & Freq & Med & Rare & Total & Freq & Med & Rare & Total &  \\
\hline
MLE & \textbf{13.30} & 146.47 & 438.67 & 15.51 & \textbf{9107} & 3945 & 91 & 13143 & 0.377 \\
AGG & 13.35 & \textbf{146.44} & \textbf{75.39} & 15.51 & 9105 & \textbf{4287} & \textbf{345} & \textbf{13737} & \textbf{0.813} \\
\hline
Human & $-$ & $-$  & $-$  & $-$  & 10844 & 7146 & 300 & 18920 & $-$  \\
\bottomrule
\end{tabular}
\caption{Experimental results for each token group in WikiText-103 language modeling task comparing MLE baseline and AGG.}
\label{table_langmo}
\end{table*}

\begin{table*}
\centering
\begin{tabular}{l||cccc|cccc|cc}
\hline
\toprule
\multirow{2}{*}{\textbf{Methods}} &
\multicolumn{4}{c|}{\textbf{PPL} $\downarrow$} &
\multicolumn{4}{c|}{\textbf{Uniq} $\uparrow$} & 
\multirow{2}{*}{$\mathbf{I(W)}$$\uparrow$}  \\
 & Freq & Med & Rare & Total & Freq & Med & Rare & Total &  \\
\hline
UL & \textbf{14.05} & \textbf{125.17} & 385.6 & \textbf{16.17} & 9527 & 4402 & 97 & 14026 & 0.396  \\
UL + AGG & 14.17 & 125.93 & \textbf{71.48} & 16.25 & \textbf{9625} & \textbf{4884} & \textbf{453} & \textbf{14962} & \textbf{0.654}  \\
\hline
Human & $-$ & $-$  & $-$  & $-$  & 10844 & 7146 & 300 & 18920 & $-$  \\
\bottomrule
\end{tabular}
\caption{Experimental results for each token group in WikiText-103 language modeling task comparing UL and UL+AGG.}
\label{table_langmo_ul}
\end{table*}

As we described in section \ref{sec3}, part (b) of Eq. \ref{eq4} should mainly be handled to solve the degeneration problem. To address part (b) of Eq. \ref{eq4}, given a context feature vector of the $i$-th position $\textbf{h}_i$, we introduce a gate vector $\textbf{g}_1\in \mathbb{R}^N$ as follows.
\begin{equation}\label{eq8}
    \begin{aligned}
        g_{1k} =\begin{cases}
 a_k / K & \text{if } v_k\in V_r, v_k \neq y_i \\ 
 1 & \text{else },
\end{cases}
    \end{aligned}
\end{equation}
where $g_{1k}$ denotes a $k$-th component of $\textbf{g}_1$. $\textbf{g}_1$ controls the degree to which rare token embeddings move away from non-rare feature vectors whose targets differ from each rare token embedding. Also, each component of $\textbf{g}_1$ is calculated based on the rarity of each rare token, $a_k$, so gradient gating for part (b) of Eq. \ref{eq4} is adaptive for each rare tokens.

Although part (c) of Eq. \ref{eq4}, which pushes embeddings away from the feature vectors whose targets are other rare tokens, is not to be seen as the cause of the degeneration problem in section \ref{sec3}, this part also induces the degeneration problem for the certain situation when rare tokens degenerate other rare tokens. To address this, we approximate the multiple levels of rarity in the rare token group to two levels in this paper: `less rare' and `very rare'. We define the two rarity levels based on the average number of appearances of the entire rare tokens: if the token appearance $a_k$ is smaller than the mean of $a_r$ where $r\in V_r$, corresponding token is a very rare token. For the very rare token embeddings, part (c) of the gradient about embeddings pushes them away from the feature vectors whose targets are less rare tokens that are relatively frequent compared to them. This means that part (c) roles like part (b) in the above situation, which becomes the cause of the degeneration problem. Therefore, we need to handle part (c) of Eq. \ref{eq4} for very rare tokens. To address part (c) of Eq. \ref{eq4} for the very rare token embeddings, we introduce another gate vector $\textbf{g}_2\in \mathbb{R}^N$ as follows.
\begin{equation}\label{eq9}
    \begin{aligned}
        g_{2k} =\begin{cases}
min(\frac{a_k}{\bar{a}_r}, 1) & \text{if } v_k \in V_r, v_k \neq y_i \\ 
1 & \text{else},
\end{cases}
    \end{aligned}
\end{equation}
where $g_{2k}$ is the $k$-th component of $\textbf{g}_2$ and $\bar{a}_r$ is the mean of $a_r$ where $r\in V_r$. $\textbf{g}_2$ controls the degree to which very rare token embeddings move away from less rare feature vectors whose targets differ from each very rare token embedding. Also, each component of $\textbf{g}_2$ is calculated based on the rarity of each very rare token, $a_k$, so gradient gating for part (c) of Eq. \ref{eq4} is adaptive for each very rare tokens.

To calculate the loss of $\textbf{h}_i$, we calculate three logits, $\textbf{z}^0_i, \textbf{z}^1_i, \text{ and } \textbf{z}^2_i$, as follows.
\begin{equation}\label{eq10}
    \begin{aligned}
        & \textbf{z}^0_i=\textbf{h}_i\tilde{\textbf{W}}^T \\
        &\textbf{z}^l_i=\textbf{g}_l\odot \tilde{\textbf{h}}_i\textbf{W}^T + (1-\textbf{g}_l)\odot \tilde{\textbf{h}}_i\tilde{\textbf{W}}^T,
    \end{aligned}
\end{equation}
where $\textbf{W}$ denotes an embedding matrix, and $l=1,2$. Because our method solely handles the gradient for embeddings, we calculate $\textbf{z}^0_i$ for a gradient about $\textbf{h}_i$, which does not need to be gated. Finally, the negative log-likelihood loss for $i$-th position $L_i$ is computed as follows.
\begin{equation}\label{eq11}
    \begin{aligned}
        L_i
        &=-\log p^0_{I(y_i)|i} \\
        &- \mathbbm{1}(y_i\notin V_r)\log p^1_{I(y_i)|i} \\
        &- \mathbbm{1}(y_i\in V_r)\log p^2_{I(y_i)|i},
    \end{aligned}
\end{equation}
where $p_{I(y_i)|i}^m$ = $[\text{softmax}(\textbf{z}_i^m)]_{I(y_i)}$ with $m$=$0,1,2$ and  $\mathbbm{1}$($\cdot$) denotes the Indicator function. 
% Gradient for rare token embeddings is computed to:
% \begin{equation}\label{eq12}
%     \begin{aligned}
%         \nabla_{\textbf{w}_r}L_i =\begin{cases}
% (p_{r|i}-1)\textbf{h}_i & \text{if } y_i=v_r \\ 
% g_{1r}p_{r|i}\textbf{h}_i & \text{if } y_i\notin V_r \\
% g_{2r}p_{r|i}\textbf{h}_i & \text{else},  
% \end{cases}
%     \end{aligned}
% \end{equation}
% where $p_{r|i}$ = $[\text{softmax}(\textbf{z}_i^m)]_r$ whose value is irrespective of $m$. Eq. \ref{eq12} demonstrates that $L_i$ passes gradients gated by $\textbf{g}_1, \textbf{g}_2$ to rare token embeddings, which is consistent with our intent. 
Derivation of the gradient for rare token embeddings, $\nabla_{\textbf{w}_r}L_i$, is provided in Appendix \ref{appenA}.

\section{Experiments}
\begin{table*}[t]
\small
\centering
\begin{tabular}{p{1.3cm} p{12cm} p{1.3cm}}
\hline
\toprule
\textbf{Method} & \textbf{Texts} & \textbf{Uniq} $\uparrow$ \\
\hline
Prefix & 
 No. 20 Squadron is a Royal Australian Air Force ( RAAF ) support squadron . Coming under the control of No. 96 Wing , it is responsible for the management of the airfield at RAAF Base Woomera , South Australia . The squadron &
\\
\hline
\hline
MLE & 
is responsible for air defence , air defence , and air defence , as well as air defence , aerial reconnaissance , and air defence . It is also responsible for air defence , air defence , and air defence , as well as air defence , aerial reconnaissance , and air defence . & 
15 \\
\hline
AGG & 
was established in October 1943 at Townsville , Queensland , under the command of Group Captain Paddy Heffernan . It was initially based at Townsville , Queensland , under the control of No. 9 Operational Group , which controlled all air bases in New South Wales . It was renamed No. 1 Mobile Fighter Sector in April 1944 . & 
48 \\
\bottomrule
\end{tabular}
\normalsize
\caption{Generated texts on the Wikitext-103 test set and uniq tokens for each texts. 50 BPE tokens are given as prefix and the models are to generate the continuation of 100 next BPE tokens.}
\label{table_gensamples}
\end{table*}

We evaluate our method on various tasks including language modeling, word similarity, and machine translation. In the language modeling task, we focus on verifying the diversity of the generated texts. We test the learning of the semantic relationships between tokens on the word similarity task. Finally, we evaluate the quality of generated texts on the machine translation task. For all the experimental results below, we adopt the state-of-the-art model architecture as a baseline to properly demonstrate the effectiveness of our method. Every detail on the experiment, such as model hyper-parameters and training configurations, regard the reproducibility are provided in Appendix \ref{appenB}.

\subsection{Language Modeling}
\noindent\textbf{Setting} We conduct experiments using WikiText-103 dataset, which is a significantly large dataset for language modeling task with approximately 103M words and 260K vocabulary size \citep{Merity2018RegularizingAO}. Texts in the dataset are preprocessed based on the byte-pair encoding\citep{Sennrich2016NeuralMT}.  We adopt the GPT-2 medium architecture\citep{Radford2019LanguageMA}, which comprises 24 Transformer decoder layers as a baseline model. Because our method is about learning token embeddings, we train the models from scratch for a maximum of 50k iterations and evaluate them based on the perplexity of the validation set. For hyper-parameter searching, we select $\alpha\in \{0.01,0.02,0.03,0.04,0.05\}$ for AGG method on the language modeling task. The hyper-parameter sensitivity for the AGG are given in Appendix \ref{appenD}.

We use three quantitative metrics to evaluate our method: Perplexity, Uniq, and $I(\textbf{W})$. Related to the likelihood of generated texts, Perplexity quantifies the prediction difficulty over the next token. Uniq \citep{Welleck2020NeuralTG} quantify the number of unique next-token predictions, measuring the token diversity. As described in section \ref{sec3}, $I(\textbf{W})$ measures the isotropy of the token embedding space. 

\noindent\textbf{Results} We present our results for the testset in Table \ref{table_langmo}. We denote the baseline method as `MLE' and our method as `AGG'. We measure Perplexity and Uniq for each token group defined in Section \ref{sec3}. As presented in Table \ref{table_langmo}, AGG improves the overall metrics for the medium and rare groups while maintaining performance for the frequent token group. This shows that our method not only improves the quality of rare token embeddings, but also the quality of non-rare token embeddings. In particular, for the rare group, the Perplexity score decrease significantly and the number of unique predictions surpasses the human distribution. The $I(\textbf{W})$ for all token embeddings increased over 2 times the baseline. Experimental results of $I(\textbf{W})$ for the embeddings of each frequency groups can be found in Appendix \ref{appenI(W)}. Table \ref{table_gensamples} shows examples of generated texts from MLE baseline and AGG. We also show additional examples of generated texts in Appendix \ref{appenF}. 

\noindent\textbf{Compatibility} Neural text degeneration problem is another problem in neural text generative models, where the model generates texts that are less likely to match human word distributions. Existing methods for this problem focus on the diversity of the generated texts by adding an auxiliary loss to the original negative log-likelihood loss \citep{Welleck2020NeuralTG}. Although \citet{Welleck2020NeuralTG} and AGG attempts to address the same problem about diversity, AGG can be compatible with the existing method in the text degeneration problem because AGG does not alter the form of the loss function in MLE training. Table \ref{table_langmo_ul} presents the results of the experiments about fusion of unlikelihood training\citep{Welleck2020NeuralTG} and AGG. We denote the unlikelihood training as UL. From Table \ref{table_langmo_ul}, we notice that when UL and AGG are fused, it produces a synergistic effect that exceeds the gain of each for the baseline. This indicates that AGG is compatible with methods that address other problems in text generation.

\subsection{Word Similarity}
\noindent\textbf{Setting} We evaluate the semantic relationship between tokens for AGG and the baseline with four word similarity datasets: MEN, WS353, RG65, and RW(\citealp{article}; \citealp{agirre-etal-2009-study}; \citealp{rg65}; \citealp{Luong-etal:conll13:morpho}). Methods are tested whether the similarity between the given two words in the embedding space is consistent with the ground truth, in terms of Spearman's rank correlation. We adopt cosine distance to compute the similarity between embeddings. We use the same models trained on language modeling tasks with the WikiText-103 dataset for the word similarity task.

\noindent\textbf{Results} Table \ref{table_wordsim} presents the result obtained from the evaluation of the word similarity task. From this table, it can be observed that our method outperforms the baseline on overall datasets. Although AGG handles only training of rare tokens, the semantic relationships between all tokens are also well learned. Qualitative studies on semantic alignment between tokens are provided in Appendix \ref{appenE}.

\begin{table}[t]
\centering
\begin{tabular}{l||c c}
\hline
\toprule
\textbf{Datasets} & \textbf{MLE} & \textbf{AGG} \\
\hline
MEN & 33.57 & \textbf{55.13}  \\
WS353 & 47.51 & \textbf{56.54} \\
RG65 & 35.48 & \textbf{65.45}  \\
RW & 32.13 & \textbf{36.36}   \\
\bottomrule
\end{tabular}
\caption{Performance(Spearman's $\gamma \times 100$) of the models on the four word similarity datasets.}
\label{table_wordsim}
\end{table}

\begin{table}[t]
\centering
\resizebox{\columnwidth}{!}{
\begin{tabular}{l||c c}
\hline
\toprule
\multirow{2}{*}{\textbf{Methods}} & \multicolumn{2}{c}{\textbf{BLEU} $\uparrow$} \\
& \texttt{Base} & \texttt{Big} \\
\hline
Transformer \citep{Vaswani2017AttentionIA} & 27.30 & 28.40 \\
CosReg \citep{Gao2019RepresentationDP}  & 28.38 & 28.94 \\
Adv MLE \citep{pmlr-v97-wang19f} & 28.43 & 29.52 \\
SC \citep{Wang2020ImprovingNL}  & 28.45 & 29.32 \\
AGG & \textbf{28.70} & \textbf{29.81} \\
\bottomrule
\end{tabular}}
\caption{Comparison of different methods in terms of BLEU scores.}
\label{table:overall_BLEU_scores}
\end{table}

% \begin{figure*}[t]
%      \centering
%      \begin{subfigure}[t]{0.32\textwidth}
%          \centering
%          \includegraphics[width=\textwidth]{sections/figs/emb_mle.pdf}
%          \caption{MLE}
%      \end{subfigure}
%      \hfill
%      \begin{subfigure}[t]{0.32\textwidth}
%          \centering
%          \includegraphics[width=\textwidth]{sections/figs/emb_agg.pdf}
%          \caption{AGG}
%      \end{subfigure}
%      \hfill
%      \begin{subfigure}[t]{0.32\textwidth}
%          \centering
%          \includegraphics[width=\textwidth]{sections/figs/singular_values.pdf}
%          \caption{Singular value decay}
%      \end{subfigure}
%      \caption{(a), (b) Token embedding visualization for the baseline model and AGG on the language modeling task with WikiText-103; (c) Normalized singular value for MLE and AGG.}
%      \label{fig_qualitative}
% \end{figure*}

\begin{figure*}[t]
     \centering
     \begin{subfigure}[t]{0.32\textwidth}
         \centering
         \includegraphics[width=\textwidth]{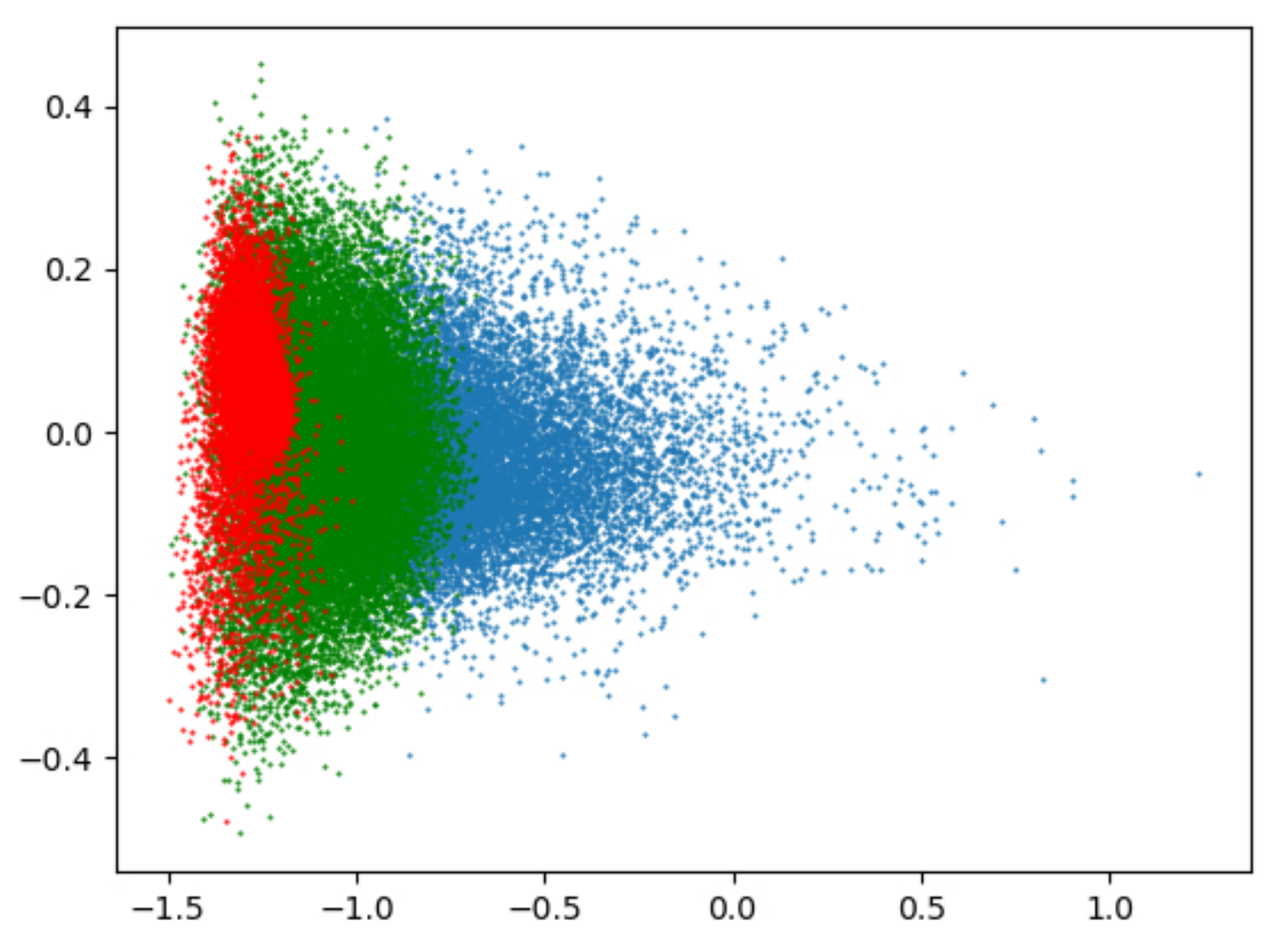}
         \caption{MLE}
     \end{subfigure}
     \hfill
     \begin{subfigure}[t]{0.32\textwidth}
         \centering
         \includegraphics[width=\textwidth]{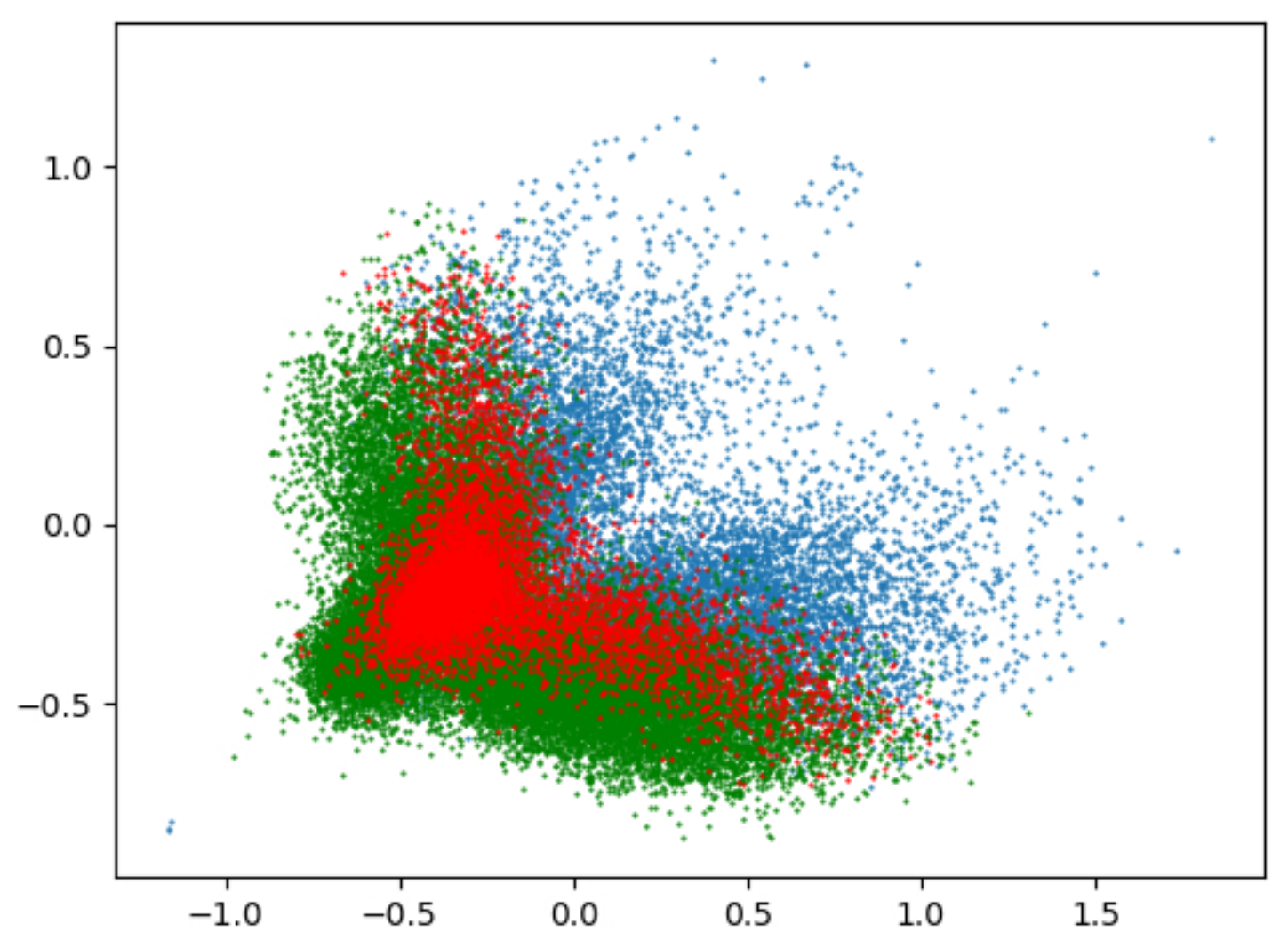}
         \caption{AGG}
     \end{subfigure}
     \hfill
     \begin{subfigure}[t]{0.32\textwidth}
         \centering
         \includegraphics[width=\textwidth]{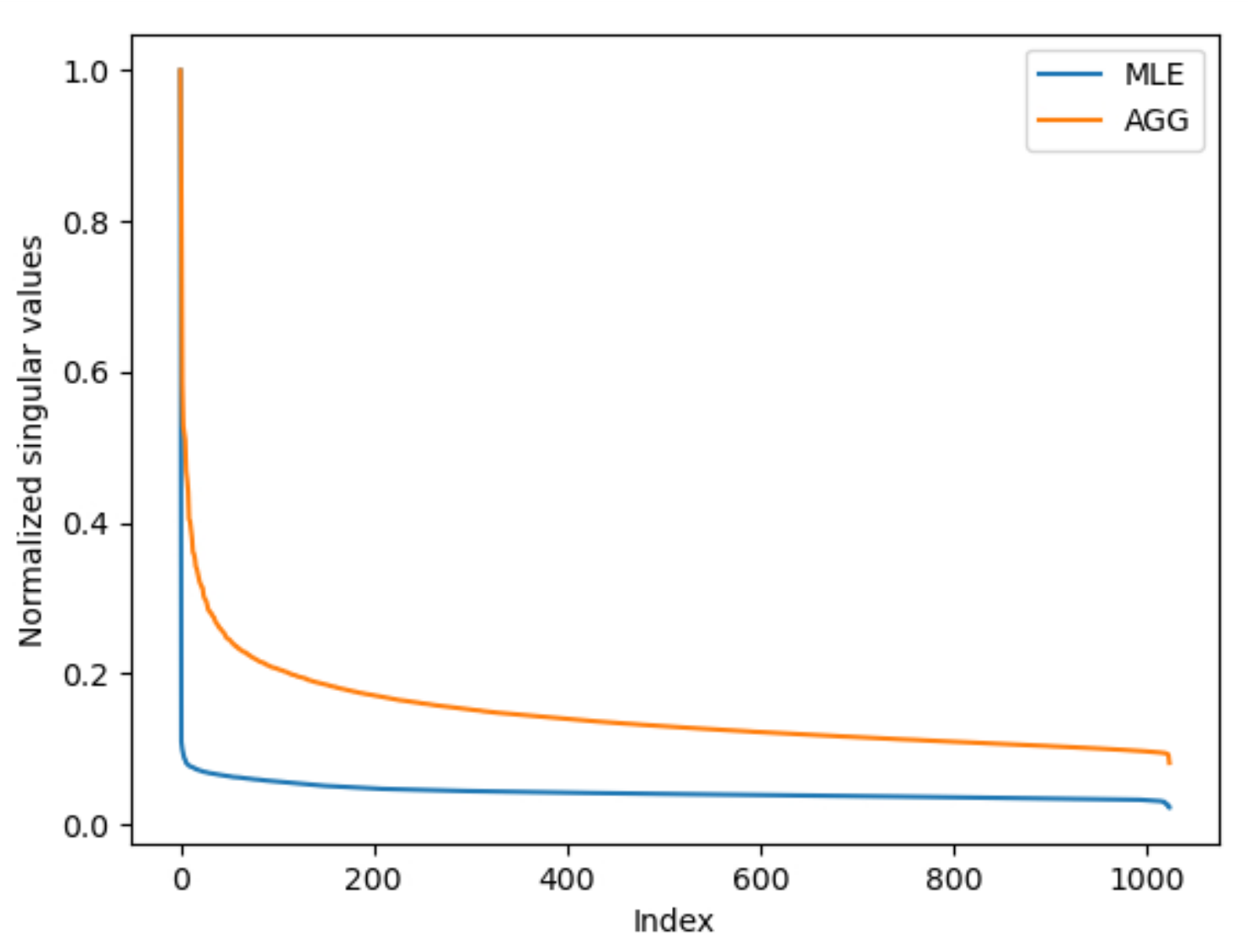}
         \caption{Singular value decay}
     \end{subfigure}
     \caption{(a), (b) Token embedding visualization for the baseline model and AGG on the language modeling task with WikiText-103. Red, green, and blue points represent rare, medium, and frequent groups respecively; (c) Normalized singular value for MLE and AGG.}
     \label{fig_qualitative}
\end{figure*}

\subsection{Machine Translation}
\noindent\textbf{Setting} We utilize a dataset from standard WMT 2014 containing 4.5M English$\rightarrow$German sentence pairs. The source and target sentences are encoded by 37K shared tokens based on byte-pair encoding\citep{Sennrich2016NeuralMT}. We adopt the two version of Transformer\citep{Vaswani2017AttentionIA} as the baseline model for applying our method: \texttt{base} and \texttt{big}. The model configuration is the same as that proposed in \citet{Vaswani2017AttentionIA}. To evaluate the quality of the generated texts, we measure BLEU score \citep{papineni-etal-2002-bleu}, which is standard metric for machine translation task.

\noindent\textbf{Results} Table \ref{table:overall_BLEU_scores} presents a comparison of our method and other methods in terms of the BLEU score. Our method achieves 1.4 and 1.41 BLEU score improvements on the machine translation task for the \texttt{base} and \texttt{big} baseline models. In addition, our method is better than all other previous works in handling the representation degeneration problem that reported BLEU scores in the same tasks. These results demonstrate the effectiveness of AGG in the quality of the generated texts. While other methods addressing the degeneration problem targets all token embeddings, target of AGG, rare token embeddings, are optimized based on the analysis about the training dynamics of token embeddings. Due to this difference, our method can prevent the over regularization problem for frequent token embeddings, which is the main advantage of AGG compared to other works. Qualitative study about cross-lingual semantic alignment between tokens of the source and target languages is provided in Appendix \ref{appenE}.

\section{Analysis of AGG}

\begin{table}
\centering
\begin{tabular}{l||c|c|c}
\hline
\toprule
\textbf{Method} & \textbf{PPL}$\downarrow$ & \textbf{Uniq}$\uparrow$ & $\mathbf{I(W)}$$\uparrow$  \\
\hline
MLE & 15.51 & 13143 & 0.377 \\
AGG & 15.51 & \textbf{13737} & \textbf{0.813} \\
no $\textbf{g}_1$ & \textbf{15.48} & 13018 & 0.367 \\
no $\textbf{g}_2$ & 15.51 & 13682 & 0.701 \\
\bottomrule
\end{tabular}
\caption{Ablation study on gating vector of AGG.}
\label{table_abl}
\end{table}

\begin{table}
\centering
\begin{tabular}{l||c|c|c}
\hline
\toprule
\textbf{Method} & \textbf{PPL}$\downarrow$ & \textbf{Uniq}$\uparrow$ & $\mathbf{I(W)}$$\uparrow$  \\
\hline
MLE & \textbf{15.51} & 13143 & 0.377 \\
AGG & \textbf{15.51} & \textbf{13737} & \textbf{0.813} \\
static AGG & 15.55 & 13614 & 0.752 \\
\bottomrule
\end{tabular}
\caption{Ablation study about dynamic grouping of AGG.}
\label{table_abl_grouping}
\end{table}

\begin{figure}[t]
     \centering
     \includegraphics[width=0.45\textwidth]{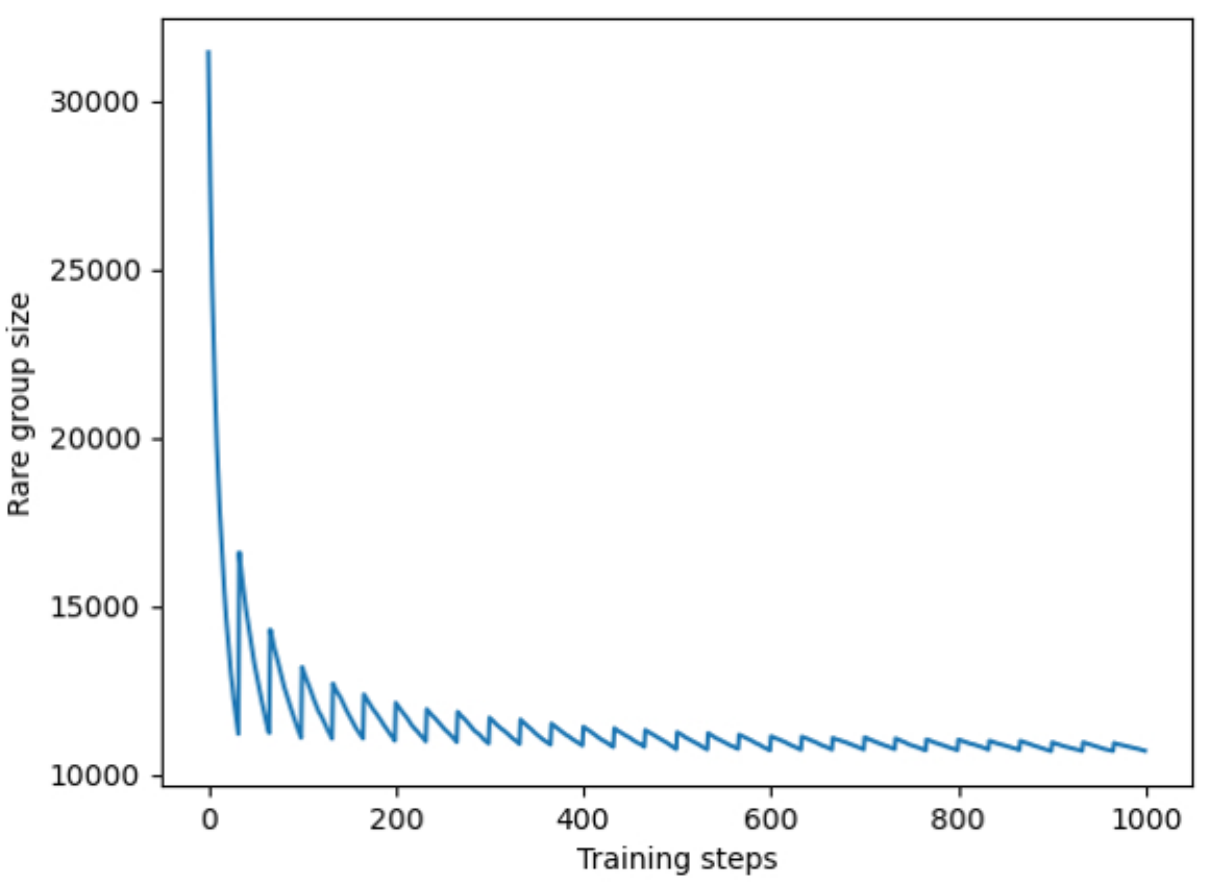}
     \caption{Size of the rare token group during initial 1k steps of training with WikiText-103 dataset.}
     \label{fig_raregroupsize}
\end{figure}

\subsection{Ablation Study}
In our method, AGG, we introduce two gate vectors, $\textbf{g}_1$, and $\textbf{g}_2$, to handle the gradient for rare and very rare token embeddings. We conduct experiments on these gate vectors. Table \ref{table_abl} presents the results of the ablation studies compared with the MLE and AGG. When $\textbf{g}_1$ is excluded from AGG (denoted as `no $\textbf{g}_1$'), Uniq and $I(\textbf{W})$ decreased significantly, because $\textbf{g}_1$ is the key component for the gradient gating. When $\textbf{g}_2$ is excluded from AGG (denoted as `no $\textbf{g}_2$'), Uniq and $I(\textbf{W})$ slightly decrease. Accordingly, we notice that $\textbf{g}_2$ is important for the gating of gradients fort the very rare token embeddings.

Also, we present the analysis about rare token grouping method of AGG. Figure \ref{fig_raregroupsize} presents the size of the rare token group during initial 1k training steps when the model is trained with WikiText-103 dataset. As presented in the figure, rare group size fluctuate wildly at the initial training stage. We expect for this grouping method to determine an optimal rare token group for the current training step. Table \ref{table_abl_grouping} presents the results of ablation study about dynamic grouping. To except dynamic grouping from AGG, we fixed the rare token group after 1 epoch. For this static grouping AGG method, Next-token diversity(Uniq) and the isotropy of the token embedding space($I(\textbf{W})$) perform worse than dynamic grouping AGG.

\subsection{Visualization}
Figure \ref{fig_qualitative} (a) and (b) present the visualizations of the embedding space of baseline MLE and our method. In the figure, applying the AGG method restores the isotropy of the token embedding space. In addition, we observe that the regions occupied by each token group are not disjoint when applying AGG. For baseline, the regions occupied by rare group and the frequent group are disjoint, which is refered as the frequency bias problem of embeddings \citep{Gong2018FRAGEFW}. From the analysis of the visualization of the embedding space, we notice that the manipulating the training of the rare token embeddings can alleviate the frequency bias problem. Figure \ref{fig_qualitative} (c) presents the plot of the normalized singular value of embedding matrix for MLE and AGG. Slowly decaying singular values of AGG demonstrate an isotropic distribution of the embedding space.

% \begin{table}[t]
% \centering
% \begin{tabular}{l||c|c|c}
% \hline
% \toprule
% \textbf{Method} & \textbf{PPL}$\downarrow$ & \textbf{Uniq}$\uparrow$ & $\mathbf{I(W)}$$\uparrow$  \\
% \hline
% MLE & 15.51 & 13143 & 0.377 \\
% AGG & 15.51 & 13737 & 0.813 \\
% no $\textbf{g}_1$ & 15.48 & 13018 & 0.367 \\
% no $\textbf{g}_2$ & 15.51 & 13682 & 0.701 \\
% \bottomrule
% \end{tabular}
% \caption{Ablation study on gating vector of AGG.}
% \label{table_abl}
% \end{table}

% \subsection{Ablation Study}
% In our method, AGG, we introduce two gate vectors, $\textbf{g}_1$, and $\textbf{g}_2$, to handle the gradient for rare and very rare token embeddings. We conduct experiments on these gate vectors. Table \ref{table_abl} presents the results of the ablation studies compared with the MLE and AGG. When $\textbf{g}_1$ is excluded from AGG (denoted as 'no $\textbf{g}_1$'), Uniq and $I(\textbf{W})$ decreased significantly, because $\textbf{g}_1$ is the key component for the gradient gating. When $\textbf{g}_2$ is excluded from AGG (denoted as 'no $\textbf{g}_2$'), Uniq and $I(\textbf{W})$ slightly decrease. Accordingly, we notice that $\textbf{g}_2$ is important for the gating of gradients fort the very rare token embeddings. The analysis of rare token grouping is also important for our study, and it can be found in Appendix \ref{appenC}.

\section{Conclusion}
In this study, we analyzed the training dynamics of the token embeddings concerning the representation degeneration problem of the learned embeddings, focusing on the rare tokens. Based on the analysis, we propose an adaptive gradient gating method that solves the problem by solely handling the training for rare token embeddings. Experiments and qualitative studies in various tasks of text generation demonstrate the effectiveness of our method. 
% By optimizing the target of the method, AGG can prevent over regularization for frequent tokens. Also, AGG is orthogonal to the existing method in the neural text degeneration problem, which means it can be compatible to fuse AGG and the existing methods in other problems. 
Beyond the two-level approximation of rarity of rare tokens which is applied to our study, addressing multiple levels of rarity can be an interesting region to study for the future work.

\section*{Acknowledgements}
 This work was supported by Institute of Information \& communications Technology Planning \& Evaluation (IITP) grant funded by the Korea government(MSIT) [NO.2021-0-01343, Artificial Intelligence Graduate School Program (Seoul National University)], the BK21 FOUR program of the Education and Research Program for Future ICT Pioneers, Seoul National University in 2022, AIRS Company in Hyundai Motor Company \& Kia Corporation through HMC/KIA-SNU AI Consortium Fund, and SNU-Naver Hyperscale AI Center.

\bibliography{acl2022}
\bibliographystyle{acl_natbib}

\begin{appendices}
\section{Derivation of the gradient of AGG loss \textit{w.r.t.} rare token embedding} \label{appenA}
We follow the same notation as in the main paper. Before we write the derivation of the gradient about rare token embedding $\textbf{w}_r$, we write the gradient of $f(\Tilde{\textbf{w}}_j)$ and $(z_i^l)_j$ about $\textbf{w}_r$, where $f(\Tilde{\textbf{w}}_j)$ is the function of $\Tilde{\textbf{w}}_j$ with $j=1, ..., N$ and $(z_i^l)_j$ is a $j$-th component of $\textbf{z}_i^l$ with $l=0,1,2$ as follows.
\begin{equation}\label{eq12}
    \begin{aligned}
        \nabla_{\textbf{w}_r}f(\Tilde{\textbf{w}}_j)
        &=\nabla_{\Tilde{\textbf{w}}_j}f(\Tilde{\textbf{w}}_j) \odot 
        \nabla_{\textbf{w}_r}\Tilde{\textbf{w}}_j \\
        &=\nabla_{\Tilde{\textbf{w}}_j}f(\Tilde{\textbf{w}}_j) \odot 0 \\
        &= 0 \text{ for all } j\\
        &(\because \Tilde{\textbf{w}}_j \text{ is treated as a constant.})
    \end{aligned}
\end{equation}
\begin{equation}\label{eq13}
    \begin{aligned}
        \nabla_{\textbf{w}_r}(z_i^l)_j
        &=\nabla_{\textbf{w}_r}[g_{lj}\cdot \Tilde{\textbf{h}}_i\textbf{w}_j^T + 
        (1 - g_{lj}\cdot \Tilde{\textbf{h}}_i\Tilde{\textbf{w}_j^T)}] \\
        &=g_{lj}\nabla_{\textbf{w}_r}\Tilde{\textbf{h}}_i\textbf{w}_j^T + 0 \\
        &=\begin{cases}
        g_{lj}\Tilde{\textbf{h}}_i & \text{if } j=r \\
        0 & \text{else}
        \end{cases} \\
        &=\begin{cases}
        g_{lj}\textbf{h}_i & \text{if } j=r \\
        0 & \text{else}
        \end{cases} \\
        &(\because \textbf{h}_i = \Tilde{\textbf{h}}_i \text{ in terms of value.})
    \end{aligned}
\end{equation}

Considering the case of $y_i \notin V_r$, AGG negative log-likelihood loss for the $i$-th position of token generation, $L_i^{AGG}$ is written as follows.
\begin{equation}\label{eq14}
    \begin{aligned}
        L_i^{AGG}=-\log p^0_{I(y_i)|i} -\log p^1_{I(y_i)|i} 
    \end{aligned}
\end{equation}

Then gradient of $L_i^{AGG}$ about $\textbf{w}_r$ is written as follows.
\begin{equation}\label{eq15}
    \begin{aligned}
        &\nabla_{\textbf{w}_r}L_i^{AGG} \\
        &=-\nabla_{\textbf{w}_r}\log p^0_{I(y_i)|i} 
        - \nabla_{\textbf{w}_r}\log p^1_{I(y_i)|i} \\
        &=-\nabla_{\textbf{w}_r}\log p^1_{I(y_i)|i} - 0 \\
        &(\because \log p^0_{I(y_i)|i} \text{ is a function of $\tilde{\textbf{w}}_r$.}) \\
        &=-\frac{1}{p^1_{I(y_i)|i}}\nabla_{\textbf{w}_r}p^1_{I(y_i)|i}\\
        &=-\frac{1}{p^1_{I(y_i)|i}}\sum_{j=1}^N\nabla_{(z_i^1)_j}p^1_{I(y_i)|i} \cdot
        \nabla_{\textbf{w}_r}(z_i^1)_j \\
        &(\because p^1_{I(y_i)|i} \text{ is a function of $(z_i^1)_j, j=1, ..., N$.}) \\
        &=-\frac{1}{p^1_{I(y_i)|i}}\nabla_{(z_i^1)_r}p^1_{I(y_i)|i} \cdot
        \nabla_{\textbf{w}_r}(z_i^1)_r \\
        &(\text{By Eq. \ref{eq13}.}) \\
    \end{aligned}
\end{equation}

As $p^1_{I(y_i)|i}=[\text{softmax}(\textbf{z}_i^1)]_{I(y_i)|i}$,
\begin{equation}\label{eq16}
    \begin{aligned}
        \nabla_{(z_i^1)_r}p^1_{I(y_i)|i}=-p^1_{I(y_i)|i}p^1_{r|i}.
    \end{aligned}
\end{equation}
Thus, $\nabla_{\textbf{w}_r}L_i^{AGG}$ is computed as follows.
\begin{equation}\label{eq17}
    \begin{aligned}
        &\nabla_{\textbf{w}_r}L_i^{AGG} \\
        &=-\frac{1}{p^1_{I(y_i)|i}}\nabla_{(z_i^1)_r}p^1_{I(y_i)|i} \cdot
        \nabla_{\textbf{w}_r}(z_i^1)_r \\
        &(\text{By Eq. \ref{eq15}.}) \\
        &=p^1_{r|i}\cdot \nabla_{\textbf{w}_r}(z_i^1)_r \\
        &=g_{1r}p^1_{r|i}\textbf{h}_i \\
        &(\text{By Eq. \ref{eq13}.})
    \end{aligned}
\end{equation}

Considering the case of $y_i \in V_r$ but $y_i \neq v_r$, $L_i^{AGG}$ is written as follows.
\begin{equation}\label{eq18}
    \begin{aligned}
        L_i^{AGG}=-\log p^0_{I(y_i)|i} -\log p^2_{I(y_i)|i} 
    \end{aligned}
\end{equation}

Then $\nabla_{\textbf{w}_r}L_i^{AGG}$ is written as follows.
\begin{equation}\label{eq19}
    \begin{aligned}
        &\nabla_{\textbf{w}_r}L_i^{AGG} \\
        &=-\nabla_{\textbf{w}_r}\log p^0_{I(y_i)|i} 
        - \nabla_{\textbf{w}_r}\log p^2_{I(y_i)|i} \\
        &=-\nabla_{\textbf{w}_r}\log p^2_{I(y_i)|i} - 0 \\
        &(\because \log p^0_{I(y_i)|i} \text{ is a function of $\tilde{\textbf{w}}_r$.}) \\
        &=-\frac{1}{p^2_{I(y_i)|i}}\nabla_{\textbf{w}_r}p^2_{I(y_i)|i}\\
        &=-\frac{1}{p^2_{I(y_i)|i}}\sum_{j=1}^N\nabla_{(z_i^2)_j}p^2_{I(y_i)|i} \cdot
        \nabla_{\textbf{w}_r}(z_i^2)_j \\
        &(\because p^2_{I(y_i)|i} \text{ is a function of $(z_i^2)_j, j=1, ..., N$.}) \\
        &=-\frac{1}{p^2_{I(y_i)|i}}\nabla_{(z_i^2)_r}p^2_{I(y_i)|i} \cdot
        \nabla_{\textbf{w}_r}(z_i^2)_r \\
        &(\because \text{Eq. \ref{eq13}.}) \\
    \end{aligned}
\end{equation}

As $p^2_{I(y_i)|i}=[\text{softmax}(\textbf{z}_i^2)]_{I(y_i)|i}$,
\begin{equation}\label{eq20}
    \begin{aligned}
        \nabla_{(z_i^2)_r}p^2_{I(y_i)|i}=-p^2_{I(y_i)|i}p^2_{r|i}.
    \end{aligned}
\end{equation}
Thus, $\nabla_{\textbf{w}_r}L_i^{AGG}$ is computed as follows.
\begin{equation}\label{eq21}
    \begin{aligned}
        &\nabla_{\textbf{w}_r}L_i^{AGG} \\
        &=-\frac{1}{p^2_{I(y_i)|i}}\nabla_{(z_i^2)_r}p^2_{I(y_i)|i} \cdot
        \nabla_{\textbf{w}_r}(z_i^2)_r \\
        &(\text{By Eq. \ref{eq19}.}) \\
        &=p^2_{r|i}\cdot \nabla_{\textbf{w}_r}(z_i^2)_r \\
        &=g_{2r}p^2_{r|i}\textbf{h}_i \\
        &(\text{By Eq. \ref{eq13}.})
    \end{aligned}
\end{equation}

Considering the remained case of $y_i=v_r$, since $y_i\in V_r$, $L_i^{AGG}$ is same as the second case, and derivation process of $\nabla_{\textbf{w}_r}L_i^{AGG}$ shares the same process with Eq. \ref{eq19}. As $I(y_i) = r$,
\begin{equation}\label{eq22}
    \begin{aligned}
        \nabla_{(z_i^2)_r}p^2_{I(y_i)|i}=p^2_{I(y_i)|i}(1 - p^2_{I(y_i)|i})
    \end{aligned}
\end{equation}
Thus, $\nabla_{\textbf{w}_r}L_i^{AGG}$ is computed as follows.
\begin{equation}\label{eq23}
    \begin{aligned}
        &\nabla_{\textbf{w}_r}L_i^{AGG} \\
        &=-\frac{1}{p^2_{I(y_i)|i}}\nabla_{(z_i^2)_r}p^2_{I(y_i)|i} \cdot
        \nabla_{\textbf{w}_r}(z_i^2)_r \\
        &(\text{By Eq. \ref{eq22}.}) \\
        &=-(1 - p^2_{I(y_i)|i})\cdot \nabla_{\textbf{w}_r}(z_i^2)_r \\
        &=-g_{2r}(1 - p^2_{I(y_i)|i})\textbf{h}_i \\
        &(\text{By Eq. \ref{eq13}.}) \\
        &=(p^2_{r|i}-1)\textbf{h}_i \\
        &(\because I(y_i)=r \text{ and } g_{2r}=1 \text{ if } I(y_i)=r.)
    \end{aligned}
\end{equation}

As $p_{r|i}=p_{r|i}^m$ with $m=0,1,2$ in terms of value, we finally write $\nabla_{\textbf{w}_r}L_i^{AGG}$ as follows.
\begin{equation}\label{eq24}
    \begin{aligned}
        \nabla_{\textbf{w}_r}L_i =\begin{cases}
(p_{r|i}-1)\textbf{h}_i & \text{if } y_i=v_r \\ 
g_{1r}p_{r|i}\textbf{h}_i & \text{if } y_i\notin V_r \\
g_{2r}p_{r|i}\textbf{h}_i & \text{else},  
\end{cases}
    \end{aligned}
\end{equation}

\section{Experimental Details} \label{appenB}
In this section, we present the details of the experiments in main page. All the experiments were conducted with a single GPU on our machine (GPU: NVIDIA A40) and from single run. For each task in the experiments, we use the same model architecture and train it with different objectives(\textit{i.e.,} MLE, AGG, UL). The hyper-parameters used for different training methods in the same task are exactly same. The detailed hyper-parameters are described in Table \ref{table_hyperparam}.

\section{Experimental Results of I(W) for each frequency groups} \label{appenI(W)}
In this section, we present the experimental results about $I(\textbf{W})$ for the embeddings of each frequency groups. Table \ref{table_i_mle} shows the $I(\textbf{W})$ comparing MLE baseline and AGG. Table \ref{table_i_ul} shows the $I(\textbf{W})$ comparing UL baseline and the fusion of UL and AGG. As presented in Table \ref{table_i_mle} and \ref{table_i_ul}, AGG improves isotropy of the embedding space for all frequency groups, indicating that our method solves the whole degeneration problem.

\begin{table}[t]
\centering
\begin{tabular}{l||ccc}
\hline
\toprule
\multirow{2}{*}{\textbf{Methods}} &
\multicolumn{3}{c}{$\mathbf{I(W)}$$\uparrow$} \\ 
 & Freq & Med & Rare \\
\hline
MLE & 0.51 & 0.33 & 0.278  \\
AGG & \textbf{0.702} & \textbf{0.714} & \textbf{0.813} \\

\bottomrule
\end{tabular}
\caption{Experimental results about $\mathbf{I(W)}$ for each token group in WikiText-103 language modeling task comparing MLE baseline and AGG.}
\label{table_i_mle}
\end{table}

\begin{table}[t]
\centering
\begin{tabular}{l||ccc}
\hline
\toprule
\multirow{2}{*}{\textbf{Methods}} &
\multicolumn{3}{c}{$\mathbf{I(W)}$$\uparrow$} \\ 
 & Freq & Med & Rare \\
\hline
UL & 0.533 & 0.351 & 0.293  \\
UL + AGG & \textbf{0.731} & \textbf{0.626} & \textbf{0.696} \\

\bottomrule
\end{tabular}
\caption{Experimental results about $\mathbf{I(W)}$ for each token group in WikiText-103 language modeling task comparing UL baseline and UL + AGG.}
\label{table_i_ul}
\end{table}
\section{Hyperparameter Sensitivity} \label{appenD}
\begin{figure*}
     \centering
     \begin{subfigure}[t]{0.32\textwidth}
         \centering
         \includegraphics[width=\textwidth]{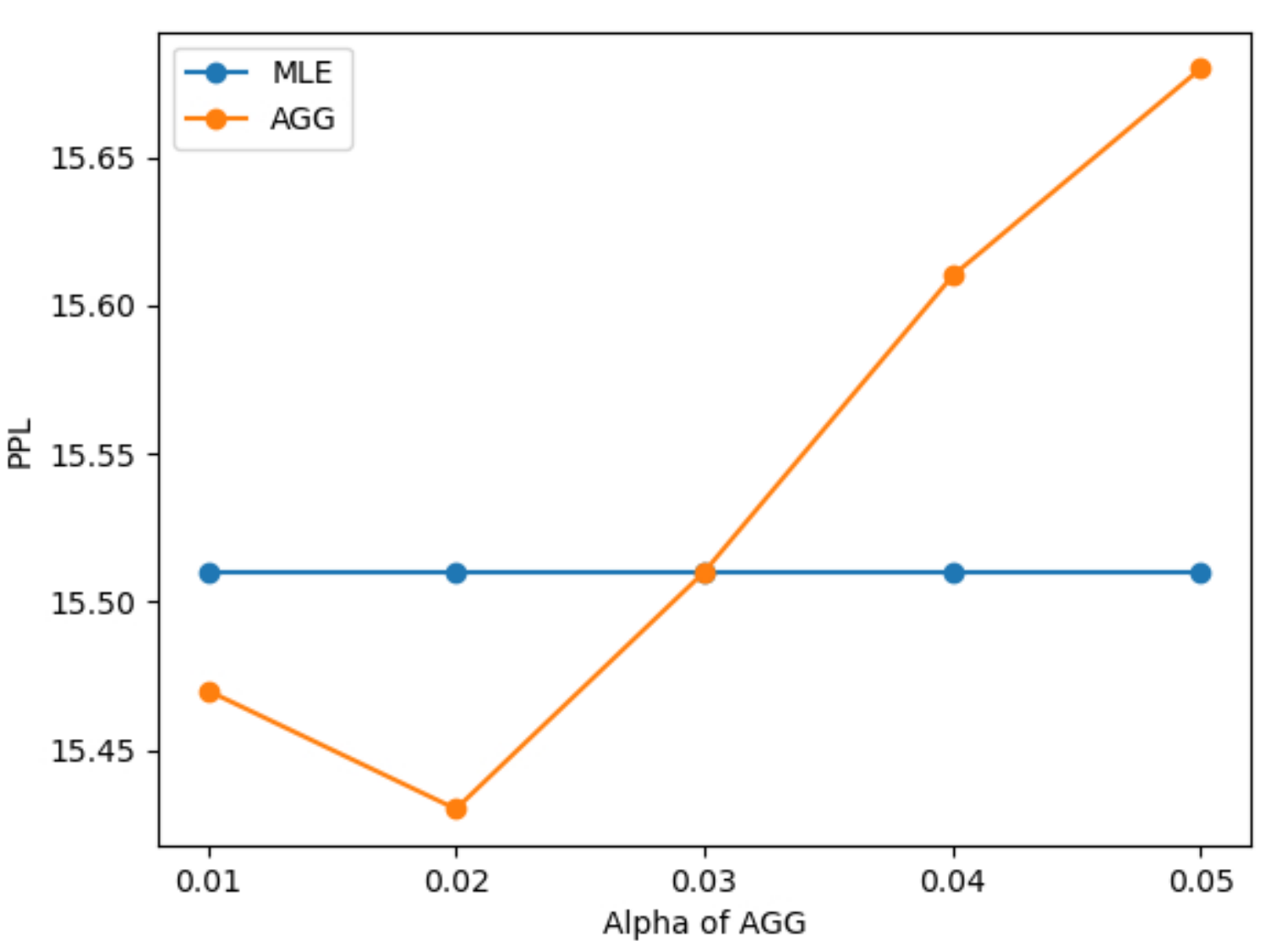}
         \caption{Perplexity}
     \end{subfigure}
     \hfill
     \begin{subfigure}[t]{0.32\textwidth}
         \centering
         \includegraphics[width=\textwidth]{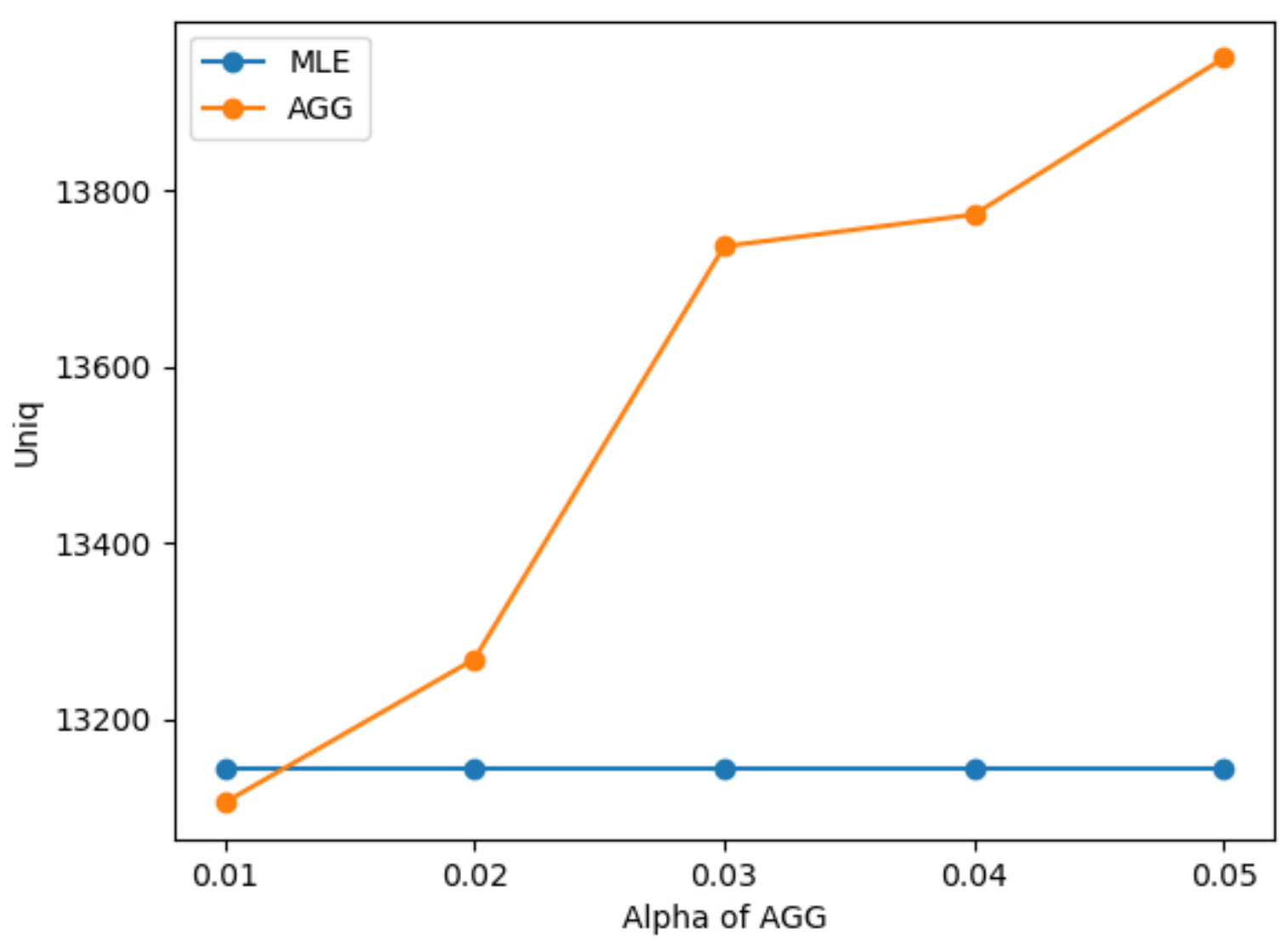}
         \caption{Uniq}
     \end{subfigure}
     \hfill
     \begin{subfigure}[t]{0.32\textwidth}
         \centering
         \includegraphics[width=\textwidth]{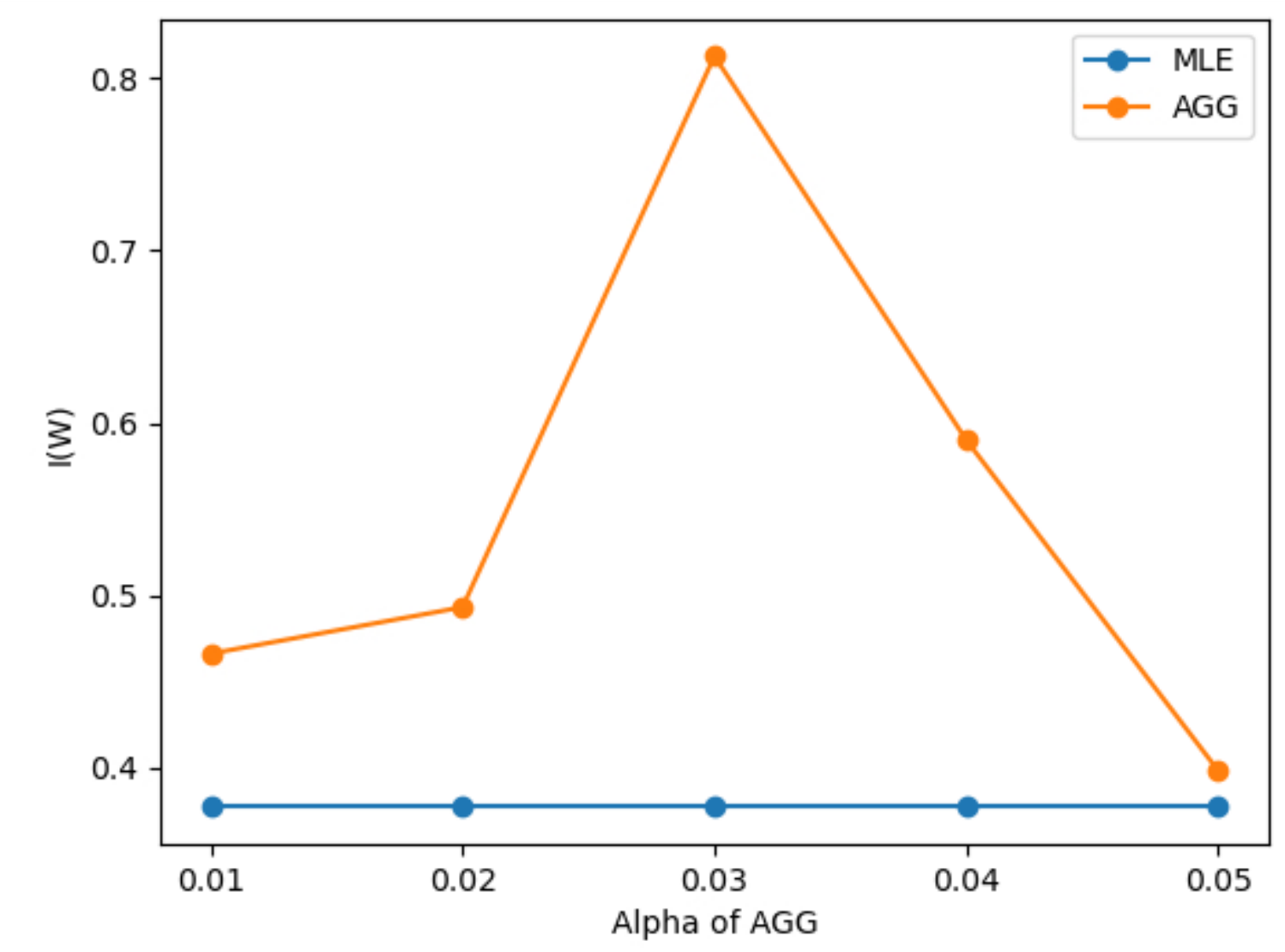}
         \caption{$I(\textbf{W})$}
     \end{subfigure}
     \caption{Hyper-parameter($\alpha$) sensitivity of AGG in the language modeling task on Wikitext-103 dataset.}
     \label{fig_hp}
\end{figure*}

In this sections we show how the metrics used on language modeling task change with the hyper-parameter $\alpha$ in Figure \ref{fig_hp}. We observed an interesting phenomenon about the non-rare token group when rare token group size increases over a specific threshold. For the rare token group, Uniq and I(W) metrics have a positive correlation. They increase together up to a certain alpha value and decrease together as alpha increases over that value. However, for the non-rare token group, Uniq increases as alpha increases over that certain value while there are negative effects where I(W) decreases and Ppl increases. Because non-rare tokens are a major group, Figure \ref{fig_hp} (b) and (c) present the above phenomenon about the non-rare token group although they present metrics for overall tokens. We consider this phenomenon to be another degeneration problem, as the increase of Uniq with negative impacts on isotropy and likelihood does not imply improvement of text quality, implying just generation of unproper tokens. This problem which occurs when rare token group size increases over a certain threshold can be handled in future work. 

\section{Qualitative Study about Semantic Alignments between Tokens} \label{appenE}
In this section, we present qualitative studies about semantic alignments between tokens for language modeling and machine translation tasks. We select three rare token from each datasets: "homepage", "Werewolf", and "policymakers" for WikiText-103 dataset, and "optimum", "criminal", and "happiness" for WMT14 En$\rightarrow$De dataset. For each rare token, we extract the top-5 nearest neighbor token predicted by the cosine distance between token embeddings. Compared with baseline MLE method, AGG shows significant improvement to train semantic alignments for rare tokens. From Table \ref{table_lm_token}, we notice that the rare tokens trained with AGG are semantically well aligned and not biased about token frequency. Table \ref{table_mt_token} demonstrates that token embeddings trained with AGG also learn the cross-lingual semantic alignments between target language tokens.

\begin{table*}
\centering
\begin{tabular}{l||c|c|cc}
\hline
\toprule
\multirow{2}{*}{\textbf{Hyperparameter}} &
\multirow{2}{*}{\textbf{Empirical Study}} &
\multirow{2}{*}{\textbf{Language Modeling}} &
\multicolumn{2}{c|}{\textbf{Machine Translation}}  \\
 &  &  & \texttt{Base} & \texttt{Big} \\
\hline
\# of layers & 6 & 24 & 6-6 & 6-6 \\
Hidden dimension & 512 & 1024 & 512 & 1024  \\
Projection dimension & 2048 & 4096 & 2048 & 4096  \\
\# of heads & 8 & 16 & 8 & 16  \\
Dropout & 0.1 & 0.1 & 0.1 & 0.3  \\
Vocabulary size & 44256 & 44256 & 40624 & 40624  \\
\# of parameters & 42M & 358M & 65M & 218M \\ 
\hline
Learning rate & $7\cdot 10^{-4}$ & $7\cdot 10^{-4}$ & $1\cdot 10^{-3}$ & $1\cdot 10^{-3}$ \\
Max tokens per batch & 32k & 32k & 64k & 64k \\
Maximum training steps & 40k & 50k & 190k & 190k \\
Warmup steps & 4k & 4k & 4k & 4k \\
Optimizer & Adam & Adam & Adam & Adam \\
Weight decay & 0.01 & 0.01 & 0.01 & 0.01 \\
$\alpha$ for AGG & $-$ & 0.03 & 0.08 & 0.08  \\ 
$\alpha$ for UL & $-$ & 1.0 & $-$ & $-$\\
\bottomrule
\end{tabular}
\caption{Model configurations and training hyper-parameters for all experiments conducted in the main page. For word similarity task, the model trained on language modeling task are evaluated for word similarity datasets.}
\label{table_hyperparam}
\end{table*}

\begin{table*}[t]
\centering
\begin{tabular}{cc|cc|cc}
\hline
\toprule
\multicolumn{2}{c|}{\textbf{homepage}}    & \multicolumn{2}{c|}{\textbf{Werewolf}}    & \multicolumn{2}{c}{\textbf{policymakers}}     \\ \hline
\multicolumn{1}{c|}{\textbf{MLE}} & \textbf{AGG}  & \multicolumn{1}{c|}{\textbf{MLE}} & \textbf{AGG} & \multicolumn{1}{c|}{\textbf{MLE}} & \textbf{AGG} \\ \hline
\multicolumn{1}{c|}{\textcolor{red}{BOX}} & website  & \multicolumn{1}{c|}{\textcolor{red}{ASUS}} & Creature & \multicolumn{1}{c|}{\textcolor{red}{Steam}} & politicians \\
\multicolumn{1}{c|}{\textcolor{red}{inbox}} & \textcolor{red}{webpage} & \multicolumn{1}{c|}{\textcolor{red}{riet}} & Nightmare  & \multicolumn{1}{c|}{\textcolor{red}{death}} & environmentalists \\
\multicolumn{1}{c|}{\textcolor{red}{livestream}} & blog & \multicolumn{1}{c|}{\textcolor{red}{480}} & Bride  & \multicolumn{1}{c|}{\textcolor{red}{Venezuel}} & activists \\
\multicolumn{1}{c|}{\textcolor{red}{namespace}} & \textcolor{red}{Tumblr}  & \multicolumn{1}{c|}{\textcolor{red}{nuclear}} & \textcolor{red}{Sneak} & \multicolumn{1}{c|}{\textcolor{red}{includ}} & planners \\
\multicolumn{1}{c|}{\textcolor{red}{hashes}} & websites & \multicolumn{1}{c|}{\textcolor{red}{ATCH}} & Sniper  & \multicolumn{1}{c|}{\textcolor{red}{\textcolor{red}{reason}}} & economists \\ 
\bottomrule
\end{tabular}
\caption{Top-5 nearest neighbors of each rare tokens in WikiText-103 dataset. Performance of AGG method is compared with the baseline MLE method. \textcolor{red}{Red} color denotes the rare tokens among neighbors.}
\label{table_lm_token}
\end{table*}

\begin{table*}
\centering
\begin{tabular}{cc|cc|cc}
\hline
\toprule
\multicolumn{2}{c|}{\textbf{optimum}}    & \multicolumn{2}{c|}{\textbf{criminal}}    & \multicolumn{2}{c}{\textbf{happiness}}     \\ \hline
\multicolumn{1}{c|}{\textbf{MLE}} & \textbf{AGG}  & \multicolumn{1}{c|}{\textbf{MLE}} & \textbf{AGG} & \multicolumn{1}{c|}{\textbf{MLE}} & \textbf{AGG} \\ \hline
\multicolumn{1}{c|}{therto} & optimal  & \multicolumn{1}{c|}{Criminal} & criminals & \multicolumn{1}{c|}{juries} & happy \\
\multicolumn{1}{c|}{ratory} & optimale$^*$ & \multicolumn{1}{c|}{criminals} & Criminal  & \multicolumn{1}{c|}{enness} & joy \\
\multicolumn{1}{c|}{consultan@@} & optimalen$^*$ & \multicolumn{1}{c|}{perpetr@@} & krimi@@$^*$  & \multicolumn{1}{c|}{ocopying} & happ@@ \\
\multicolumn{1}{c|}{sofar} & maximum  & \multicolumn{1}{c|}{secution} & kriminellen$^*$ & \multicolumn{1}{c|}{ratory} & Glück$^*$ \\
\multicolumn{1}{c|}{protection@@} & Optim@@ & \multicolumn{1}{c|}{xious} & crime  & \multicolumn{1}{c|}{sacri@@} & pleasure \\ 
\bottomrule
\end{tabular}
\caption{Top-5 nearest neighbors of each rare source tokens in WMT14 En$\rightarrow$De dataset. Performance of AGG method is compared with the baseline MLE method. The symbol @@ stands for sub-word tokenization of the dataset. The symbol * denotes the synonym token of the target language.}
\label{table_mt_token}
\end{table*}
\section{Examples} \label{appenF}
We present additional generated text samples from the model trained on language modeling task in Table \ref{table_gensamples_appen}. From the table, we notice that the model trained with AGG generates more diverse and high quality text than the baseline.

\clearpage

\begin{table*}[t]
\small
\centering
\begin{tabular}{p{1.3cm} p{12cm} p{1.3cm}}
\hline
\toprule
\textbf{Method} & \textbf{Texts} & \textbf{Uniq} $\uparrow$ \\
\hline
Prefix & 
A Company , 2nd Engineer Combat Battalion , moved to the south side of the Yongsan - Naktong River road ; D Company of the 2nd Engineer Battalion was on the north side of the road . Approximately 2 miles ( 3  &
\\
\hline
\hline
MLE & 
. 2 km ) north of the river , the 2nd Engineer Combat Battalion was on the south side of the road . The 2nd Engineer Combat Battalion was on the south side of the road .
 & 
22 \\
\hline
AGG & 
. 2 km ) north of the river , the 2nd Engineer Combat Battalion was attacked by a small force of North Korean People 's Volunteer Army ( NK ) troops . The North Koreans were able to penetrate the UN lines and capture the town . & 
43 \\
\bottomrule
\end{tabular}
\normalsize
\end{table*}

\begin{table*}[t]
\small
\centering
\begin{tabular}{p{1.3cm} p{12cm} p{1.3cm}}
\hline
\toprule
\textbf{Method} & \textbf{Texts} & \textbf{Uniq} $\uparrow$ \\
\hline
Prefix & 
 The twelve 45 - calibre 35 . 6 cm ( 14 . 0 in ) Type 41 guns of the Ise class were mounted in three pairs of twin - gun , superfiring turrets . Numbered one through &
\\
\hline
\hline
MLE & 
two , the turrets were placed in superfiring pairs , one ahead of the other . The turrets were placed in superfiring pairs , with the turrets superfiring over the wing turrets . The turrets were placed in superfiring pairs , with the turrets superfiring over the wing turrets . The turrets were placed in superfiring pairs , with the turrets superfiring over the wing turrets . The turrets were placed in superfiring pairs , with the turrets superfiring over the wing turrets . The turrets were placed in superfiring & 
19 \\
\hline
AGG & 
five from front to rear , the guns had a firing arc of 260 ° . They fired a 45 . 3 kg ( 99 lb ) projectile at a muzzle velocity of 835 m / s ( 2 , 740 ft / s ) to a range of 16 , 000 m ( 52 , 000 yd ) . They had a maximum elevation of 30 degrees and fired a 36 . 3 kg ( 77 lb ) projectile at a muzzle velocity of 835 & 
55 \\
\bottomrule
\end{tabular}
\normalsize
\end{table*}

\begin{table*}[t]
\small
\centering
\begin{tabular}{p{1.3cm} p{12cm} p{1.3cm}}
\hline
\toprule
\textbf{Method} & \textbf{Texts} & \textbf{Uniq} $\uparrow$ \\
\hline
Prefix & 
Despite the expensive reconstructions , both vessels were considered obsolete by the eve of the Pacific War , and neither saw significant action in the early years of the war . Following the loss of most of the IJN 's large aircraft carriers during the Battle  &
\\
\hline
\hline
MLE & 
of Midway , the IJN was forced to rely on the United States Navy for aircraft carriers and aircraft carriers to support operations in the Pacific . The IJN 's aircraft carriers were the first to be equipped with the new Mark 4 torpedo and the IJN 's aircraft carriers were the first to be equipped with the Mark 4 torpedo . The IJN 's aircraft carriers were the first to be equipped with the Mark 4 torpedo , and the IJN 's aircraft carriers
 & 
37 \\
\hline
AGG & 
of Midway , the IJN decided to modernize its fleet and modernize its fleet . The IJN 's new ships were designed to be capable of operating at speeds of up to 30 knots ( 56 km / h ; 35 mph ) , and were fitted with a new bow section . The ships were designed to carry 1 , 000 t ( 980 long tons ; 1 , 100 short tons ) of fuel oil , and were fitted with a pair of aircraft catap & 
63 \\
\bottomrule
\end{tabular}
\normalsize
\end{table*}

\begin{table*}[t]
\small
\centering
\begin{tabular}{p{1.3cm} p{12cm} p{1.3cm}}
\hline
\toprule
\textbf{Method} & \textbf{Texts} & \textbf{Uniq} $\uparrow$ \\
\hline
Prefix & 
 Amos was born in Macclesfield , Cheshire and was a member of the Crewe Alexandra academy until he was released at the age of 10 . While at Crewe , he also played for another local team , Bollington United , as a centre &
\\
\hline
\hline
MLE & 
- back . He was a member of the team that won the FA Youth Cup in 1989 , and was a member of the team that won the FA Youth Cup in 1990 . He was a member of the team that won the FA Youth Cup in 1990 , and was a member of the team that won the FA Youth Cup in 1992 . & 
24 \\
\hline
AGG & 
- back . He was signed by Crewe Alexandra in July 2006 , and made his debut for the club in a 2 - 1 win over Rotherham United in the League Cup on 18 August 2006 . He was loaned out to Rotherham for the rest of the 2006 - 07 season , before being released at the end of the season . & 
52 \\
\bottomrule
\end{tabular}
\normalsize
\caption{Generated texts on the Wikitext-103 test set and uniq tokens for each texts. 50 bpe tokens are given as prefix and the models are to generate the continuation of 100 next bpe tokens.}
\label{table_gensamples_appen}
\end{table*}
\end{appendices}

\end{document}